\documentclass{article}

\usepackage{microtype}
\usepackage{graphicx}
\usepackage{subcaption}
\usepackage{booktabs} % for professional tables
\usepackage{adjustbox}
\usepackage{algpseudocode}

\usepackage{hyperref}

\usepackage[accepted]{icml2025}

\usepackage{amsmath}
\usepackage{amssymb}
\usepackage{mathtools}
\usepackage{amsthm}
\usepackage{booktabs}
\usepackage[capitalize,noabbrev]{cleveref}

\theoremstyle{plain}

\theoremstyle{definition}

\theoremstyle{remark}

\usepackage[textsize=tiny]{todonotes}

\icmltitlerunning{S2TX: cross-attention Multi-Scale State-Space Transformer for Time Series Forecasting}

\begin{document}

\twocolumn[
\icmltitle{S2TX: Cross-Attention Multi-Scale State-Space Transformer for Time Series Forecasting}



\icmlsetsymbol{equal}{*}

\begin{icmlauthorlist}
%\icmlauthor{Anonymous Author}{}
\icmlauthor{Zihao Wu}{yyy}
\icmlauthor{Juncheng Dong}{yyy,equal}
\icmlauthor{Haoming Yang}{yyy,equal}
\icmlauthor{Vahid Tarokh}{yyy}
% \icmlauthor{}{sch}
% \icmlauthor{}{sch}
\end{icmlauthorlist}

\icmlaffiliation{yyy}{Department of Electrical and
Computer Engineering, Duke University, Durham, NC 27708, USA}
% \icmlcorrespondingauthor{Anonymous Author}{}
\icmlcorrespondingauthor{Zihao Wu}{zihao.wu@duke.edu}
\icmlkeywords{Machine Learning, ICML}
\vskip 0.3in
]


\printAffiliationsAndNotice{\icmlEqualContribution} % otherwise use the standard text.

\begin{abstract}

Time series forecasting has recently achieved significant progress with multi-scale models to address the heterogeneity between long and short range patterns. 
Despite their state-of-the-art performance, we identify two potential areas for improvement. 
First, the variates of the multivariate time series are processed independently. Moreover, the multi-scale (long and short range) representations are learned separately by two independent models without communication. In light of these concerns, we propose \emph{State Space Transformer with cross-attention} (S2TX). S2TX employs a cross-attention mechanism to integrate a Mamba model for extracting long-range cross-variate context and a Transformer model with local window attention to capture short-range representations. By cross-attending to the global context, the Transformer model further facilitates variate-level interactions as well as local/global communications. Comprehensive experiments on seven classic long-short range time-series forecasting benchmark datasets demonstrate that S2TX can achieve highly robust SOTA results while maintaining a low memory footprint.
\end{abstract}

\section{Introduction}
Forecasting multivariate time series represents a core learning paradigm designed to predict upcoming time steps using historical data. This machine learning task finds application across a range of domains including the economy \citep{koop2010bayesian}, epidemiology \citep{nguyen2021forecasting}, and meteorology \citep{angryk2020multivariate}. Due to its significant influence, multivariate time series forecasting has garnered considerable focus. State-of-the-art (SOTA) methods for multivariate time series forecasting predominantly utilize two types of sequence models: transformers and state-space models \cite{vaswani2017attention, gu2023mamba}. By employing these foundational structures, researchers aim to advance this research domain by harnessing two key features of multivariate time series: 1) identifying temporal dependencies and 2) understanding inter-variate correlations. Effectively integrating both temporal dynamics and the interactions between variates within a single learning model is essential for the precise forecasting of these intricate multivariate time series \citep{box2015time}. %\JD{citation?}.

%Significant progress has been made in multivariate forecasting with the introduction of the transformer architecture and the attention mechanism. 

%Another line of research, state-space models, has greatly improved the capabilities of deep learning models to retain long-range context. 

\begin{figure}[t]
    \centering
    \includegraphics[width=\linewidth]{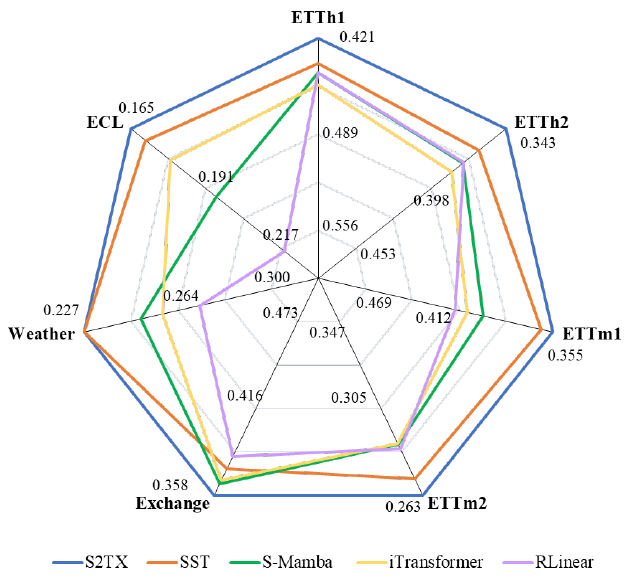} % Change to your image file
    \caption{Overview of the performance of different architectures over 7 different benchmark datasets. Average results (MSE) are reported. }
    \label{fig:overview}
\end{figure}

A recent advancement~\citet{xusst} integrates transformers and state-space models within a multi-scale framework: it first breaks down the input time series into shorter high resolution \emph{patches} and longer low resolution patches. Subsequently, it feeds the high resolution patches into a transformer model with local-attention to extract fine-grained local features and the low resolution longer patches to a state-space model (i.e., Mamba~\cite{gu2023mamba}) to learn long-range global features.
% a sequence is broken down into shorter high resolution time-series patches and fed to a transformer to leverage its ability to extract fine-grained local features, while the low resolution longer patches are inputted to the state-space model to learn long-range global features. 
This multi-scale mixture of Mamba and transformer models greatly improves the modeling of temporal dependencies. However, it leaves a crucial aspect of multivariate time series forecasting unattended, that is, the correlation between variates. 
Additionally, the local and global features are modeled independently, which overlooks the interplay between global and local features. Such global-local interplay is manifested in many real-world scenarios. For example, the commonly known \emph{El Ninõ} effect is a global, long-term weathering effect in the time-scale of years; but this global weather pattern will greatly affect the short-term local time series within days \citep{hsieh2004nonlinear}. 

\begin{figure}[t]
    \centering
    \includegraphics[width=0.8\linewidth]{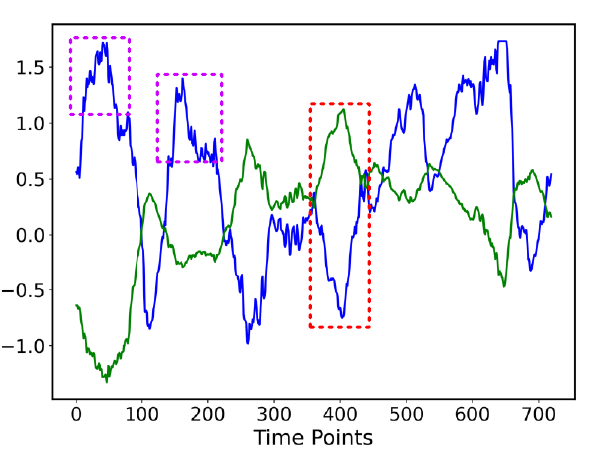} % Change to your image file
    \caption{A snippet of the weather dataset. Two variables (blue and green) were plotted over 720 time steps. The purple boxed region indicates where a global-local interaction exists, and the red boxed region indicates a cross-variate correlation. }
    \label{fig:cross-all-correlation}
\end{figure}

The cross-variate correlation and global-local features interplay, illustrated in Figure \ref{fig:cross-all-correlation}, are two crucial aspects of multivariate time series forecasting. Global patterns encompassed in the purple boxes consistently suggest increased local variation while the red-boxed region indicates a strongly inversed correlation between the two variates. To explicitly include these two crucial aspects, we introduce \emph{State Space Transformer with Cross-attention} (S2TX) where we connect cross-variate global features with fine grained local features through a carefully designed cross-attention mechanism. Specifically, we apply Mamba as the global model to process long-range, low-resolution patches across all variates in a single sweep, extracting cross-variate global context. This global context is then provided as the key and value for cross-attention to a decoder-like transformer model focusing on local, high-resolution, variate-independent patches.

\textbf{Contributions.} Our contributions are summarized below:
\begin{itemize}
    \item We identify two crucial aspects, cross-variate correlation and global-local interaction,  to improve the SOTA time series forecasting model.
    \item We propose a novel multi-scale architecture that incorporates these considerations through a cross-attention mechanism. In particular, our architecture learns variate-level correlation while leveraging the enhanced temporal learning of patchification. 
    \item We develop a cross-attention mixture of experts, enabling global-local feature interplay between a global feature-focused state-space model and a local feature-focused transformer model.
    \item We verify the efficacy of our proposed architecture on a comprehensive set of time series forecasting benchmarks.
\end{itemize}
%Recent studies have shown that processing patched time series data is more effective than elementwise processing in capturing temporal dependencies, as it introduces an inductive bias that aligns with the localized nature of time series data. Furthermore, it has been observed that using multi-scale patches enables models to learn distinct global and local features. 

%In many real-world scenarios, as illustrated in the figure, the significance lies not only in the features themselves but also in the interplay between global and local features. However, existing structures often lack an explicit mechanism for modeling this correlation. To address this, we propose using a cross-attention mechanism to explicitly model the interaction between global and local features. 

%While patch technique facilitates the capture of temporal dependencies, it transforms the time series of each variable into a multivariate time series, making it more challenging to effectively model interactions between variables. To address this, we propose leveraging Mamba as the global model to process long-range patches across all variables in a single sweep, extracting cross-variable global context. This global context is then provided as the key and value to the decoder-like transformer local model within a cross-attention mechanism. By utilizing Mamba as the global model, we take advantage of its linear time complexity and its capability to handle long series, even when the dimensionality of the data is high.

\section{Related Works}
The field of time series forecasting has seen significant evolution over the decades: shifting from classical mathematical tools \citep{bloomfield2004fourier, durbin2012time} and statistical techniques like ARIMA \citep{nerlove1971time, hyndman2018forecasting} to more recent deep learning approaches such as recurrent neural networks \citep{graves2013speech} and long-short term memory models \citep{gers2000learning}. Notably, in recent years, transformers \citep{vaswani2017attention} have demonstrated particularly promising performance on sequence modeling tasks, especially in natural language processing. Interestingly, studies have revealed that even simple linear layers can outperform complex transformer-based models in both performance and efficiency for time series forecasting \citep{zeng2023transformers, yang2024neural}.

% Making the subsections shorter so the related work can contain more paper
\textbf{Inverted Dimension.}
In investigating why transformers underperform in time series forecasting, \citet{liu2023itransformer} argues that the direct application of transformers that embed all variates is undesirable. This embedding compresses variates with distinct physical meanings and inconsistent measurement at each time step to a single token, erasing the important multivariate correlations. To address this limitation, the authors propose inverting the dimension of time and variates in the data while preserving the core mechanisms of the transformer. Many subsequent studies \citep{wang2025mamba, ahamed2024timemachine, xusst} build upon this paradigm, achieving improvements in both performance and efficiency. %This innovative approach effectively unlocks the potential of transformers for time series forecasting.
% Building on this innovative paradigm, subsequent works further refined the approach by replacing the Transformer with Mamba-based models, a promising alternative that offers linear training complexity.

\textbf{Patchification.}
Patchification of inverted data transforms the time series of each variate into a multivariate time sequence where the patches are stacked to construct an additional dimension. While patchification facilitates the capture of temporal dependencies by introducing an inductive bias aligned with the localized nature of time series data, it also overlooks the between-variate correlations due to the additional dimension: existing approaches, such as SST \citep{xusst} and PatchTST \citep{nie2022time}, treat each variate independently. Despite their strong performance, these methods lack any form of inter-variate communication. 
% Taking this out for now, maybe we can use this in 4.2? 
%Other methods like MOIRAI flatten all variables into a single sequence before patchification, but the extended sequence length, particularly in high-dimension, in addition to the quadratic complexity of the transformer, imposes a heavy computation burden.

\textbf{Mixture of Experts.}
The mixture of experts method receives increasing attention in sequence modeling after the release of Mamba \citep{gu2023mamba}. Combining the linear complexity of Mamba and the strong performance of transformers could lead to efficient and accurate sequence models. For instance, Jamba \citep{lieber2024jamba} employs a layerwise stacking of Mamba and attention layers, achieving superior performance in natural language processing compared to either component individually. For time series forecasting, SST \citep{xusst} utilizes Mamba to capture global patterns with prolonged patch lengths, while leveraging transformer to learn local details with shorter patch lengths. However, global and local patches are processed separately through each expert before their output embeddings are concatenated. Such inadequate communication between global and local features limits the integration results, restricting the model's ability to fully exploit each expert's complementary strength. 

\begin{figure*}[t]
    \centering
    \includegraphics[width=1\textwidth]{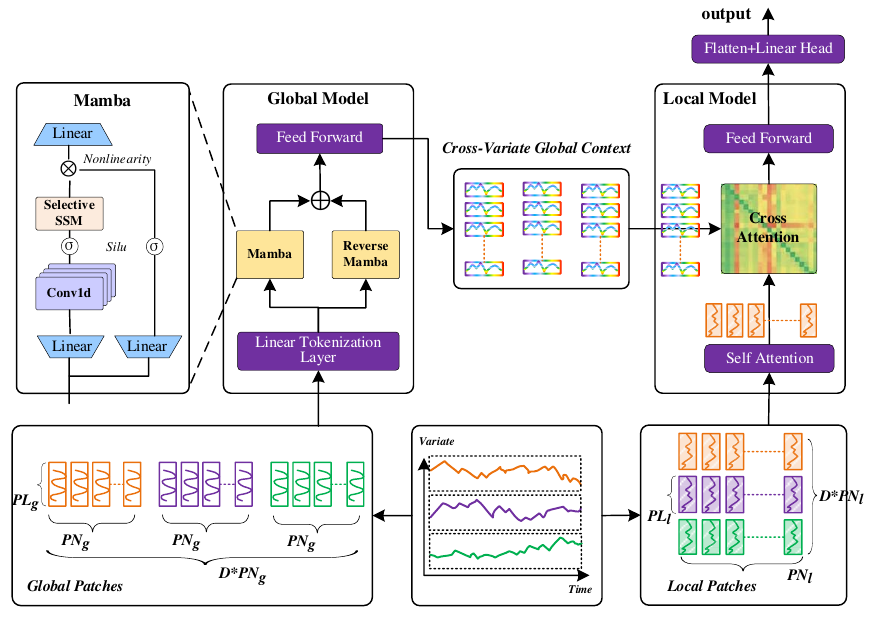} % Change to your image file
    \caption{Overview of the proposed architecture S2TX. Different variables (in different colors) of the time series are patched into global and local patches. The global patches are processed by the global model, which outputs the global context that is used to compute the key and value matrices during cross-attention with the local model. Skip connections and normalization layers are omitted for clarity of presentation.}
    \label{fig:structure}
\end{figure*}
%\textbf{Our Approach}
%\HM{Maybe we can take this out, seems redundant.}
%As we clarified the different aspects of multivariate time series forecasting, the goal is clear: with S2TX, we develop an architecture that combines the advantages of all three approaches. S2TX is a low memory architecture that maintains the learning of multivariate correlation while enabling global-local feature interplay through a cross-attentional mixture of experts.

\section{Preliminary}
\label{sec:preliminary}
In this section, we first formalize the modeling problem, then introduce the two main modules of our proposed architecture: state-space models and cross-attention. 

\subsection{Problem Setup}
\label{sec:problemSetup}
We consider the standard problem setup for time series forecasting framework \citep{liu2023itransformer}. Given a $D$-dimensional multivariate time series of length $L$ (look-back window) $\textbf{X}\in \mathbb{R}^{D\times L}$,  the goal is to predict $\textbf{Y}\in \mathbb{R}^{D\times H}$, the same $D$-dimensional multivariate time series in the future $H$ steps (horizon length). Assuming we have access to a training dataset with $N$ observations $\{\textbf{X}^{(i)},\textbf{Y}^{(i)}\}_{i=1}^N$, our goal is to learn a function $f_\phi(\textbf{X}^{(i)}): \mathbb{R}^{D\times L} \rightarrow \mathbb{R}^{D\times H}$ with parameter $\phi$ such that the mean squared error loss is minimized:
\begin{align}
    \mathcal{L}_{\mathrm{train}} = \frac{1}{N}\sum_{n=1}^{N}\|f_\phi(\textbf{X}^{(i)}) - \textbf{Y}^{(i)}\|_F^2,
\end{align}
where $F$ denotes the Frobenius norm~\citep{horn2012matrix}.

\subsection{State-Space Models}
\label{sec:SSM}
State-Space Models (SSMs) \citep{gu2020hippo, gu2021efficiently} are a family of sequence models inspired by continuous control systems described by the following equations
\begin{align}
    \mathrm{d}h = Ah + Bx,\ z = Ch + Dx,
\end{align}
where $x\in \mathbb{R}$ represents a one-dimensional input, $h\in \mathbb{R}^{d\times 1}$ is the hidden state, $z$ is the model output, $A\in \mathbb{R}^{d\times d}$, $B\in \mathbb{R}^{d\times 1}$, $C\in \mathbb{R}^{1\times d}$, and $D\in \mathbb{R}^{1\times 1}$ are parameter matrices. Matrix $D$ acts as a skip connection and is typically omitted in derivations. For multi-dimensional inputs, a stack of SSMs is employed.
The continuous system is then discretized into 
\begin{align}
    h_{t+1} = \bar{A}h_t + \bar{B}x_t,\ z_{t+1} = \bar{C}h_t,
\end{align}
where the discretized matrices are obtained with a discretization rule and a step size $\Delta$. For example, Mamba \citep{gu2023mamba} uses Zero-Order Holder rule such that $\bar{A} = \exp(\Delta A),\ 
\bar{B} = \exp(\Delta A)^{-1} (\exp(\Delta A) - \mathbb{I}) \cdot \Delta B$.

The discretized state-space models can be interpreted either as a convolutional neural network, enabling linear-time parallel training, or as a linear recurrent neural network, supporting constant-time inference, as demonstrated in S4 \citep{gu2021efficiently}. Building upon S4, Mamba extends this approach by making the matrices $B$ and $C$ input-dependent, transforming them into a selective SSM. Additionally, Mamba introduces a parallel scan algorithm to achieve linear-time training complexity.

\subsection{Cross-attention}
\label{sec:crossAttentioin}
Cross-attention is a generalization \citep{bahdanau2014neural} of self-attention \citep{vaswani2017attention}. 
% It was initially used in the decoder part of the transformer and later extended to facilitate cross-model interaction. 
Given source data $S\in \mathbb{R}^{L_S\times d_{\mathrm{model}}}$ and target data $T\in \mathbb{R}^{L_T\times d_\text{model}}$, the output of cross-attention is
\begin{align}
    \text{CrossAttention}(S,T) = \frac{(TW_q)(SW_k)^T}{\sqrt{d_\text{model}}}SW_v %\in \mathbb{R}^{L_T \times d_{\text{model}}},
\end{align}
where $W_k$, $W_q$, $W_v\in \mathbb{R}^{d_\text{model}\times d_\text{model}}$ are learnable parameters. From this perspective, self-attention can be achieved by substituting all instances of $S$ with $T$ in Cross-attention:
\begin{align}
    \text{SelfAttention}(T) = \text{CrossAttention}(T,T) %\in \mathbb{R}^{L_T \times d_{\mathrm{model}}}.
\end{align}
This cross-attention mechanism will allow us to compute cross-attentional weight where the influence of global patterns is weighted to predict a specific local pattern, allowing global-local interaction during inference. Notably, our application of cross-attention mechanism to integrate multi-scale features are commonly applied in computer vision tasks \citep{chen2021crossvit}.

\section{State-Space Transformer with Cross-attention}
\label{sec:proposedApproach}
Here we describe our proposed method State-Space Transformer with Cross-attention (S2TX). We first introduce the Multi-Scale patching process that decompose a long time series into global and local patches of different time scales. The low-resolution global patches were then fed to a Mamba-based global feature extractor to obtain the cross-variate global context. The global context is then applied as the key and value for a novel global-local cross-attention to improve the extraction of local features. Finally, we conduct a computation complexity analysis, showcasing that S2TX, with the addition of our novel cross-attention, maintained a low-memory footprint during training and inference. The general structure of S2TX is provided in Figure \ref{fig:structure}.

\begin{figure}[thbp]
    \centering
    \includegraphics[width=0.45\textwidth]{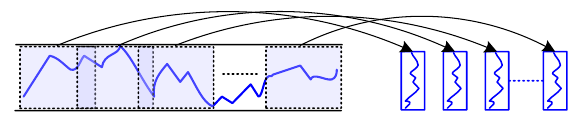} % Change to your image file
    \caption{Patch transforms a one-dimensional sequence to a sequence of patches.}
    \label{fig:patch}
\end{figure}

\subsection{Multi-Scale Patch}
\label{sec:multiscalePatch}
The patching technique has become increasingly popular for time series forecasting~\citep{gong2023patchmixer, nie2022time,xusst}. It aggregates local information into patches and effectively enhances the receptive field. Denote the sequence length of the look-back window by $L$, patch length by $PL$, stride by $STR$, and patch number by $PN$, where 
\begin{align}\label{eq:pn}
    PN = \left\lceil\frac{L-PL}{STR}\right\rceil. 
\end{align}
The patching technique transforms each (one-dimensional) variate of length $L$ into a $PL$-dimensional time series of length $PN$. More specifically, the input time series $\textbf{X}\in \mathbb{R}^{D\times L}$ is patched into $\mathbf{\tilde{X}}\in \mathbb{R}^{D\times PN\times PL}$. 

Intuitively the longer the stride, or the longer the patch length, the more long range temporal context is stored in a patch and vice versa. Utilizing this intuition, we apply the patching process onto the time series \emph{twice}: (i) one of them focuses on coarser granularity for global context, 
%Moreover, SST highlights that global patterns are more discernible at a coarser granularity, while local variations are revealed at a finer granularity. Following a similar approach, 
employing the full look-back window of length $L$, a larger patch length $PL_g$ and longer stride, along with the corresponding patch number $PN_g$ to obtain long-range global time series patches; (ii) the other leverages finer granularity with a fixed shorter look-back window of length $S$, a smaller patch length $PL_l$, and shorter stride with corresponding patch number $PN_l$ to obtain short-range local patches. The resulting two multi-scale time series patches $\mathbf{\tilde{X}}_g\in \mathbb{R}^{D\times PN_g\times PL_g}$ and $\mathbf{\tilde{X}}_l\in \mathbb{R}^{D\times PN_l\times PL_l}$ serve as inputs for the global and local models, respectively.
\begin{table*}[t]
\footnotesize
\centering
\renewcommand{\arraystretch}{0.9} % Adjust row height
\setlength{\tabcolsep}{3pt} % Adjust column spacing
\adjustbox{max width=\textwidth}{
\begin{tabular}{%lllllllllllllllllllllll
                lcccccccccccccccccccccc}
\toprule
 & \multicolumn{2}{c}{\textbf{S2TX}} & \multicolumn{2}{c}{\textbf{SST}} & \multicolumn{2}{c}{\textbf{S-Mamba}} & \multicolumn{2}{c}{\textbf{TimeM}} & \multicolumn{2}{c}{\textbf{iTrans}} & \multicolumn{2}{c}{\textbf{RLinear}} & \multicolumn{2}{c}{\textbf{PatchTST}} & \multicolumn{2}{c}{\textbf{CrossF}} & \multicolumn{2}{c}{\textbf{TimesNet}}  \\
 & \multicolumn{2}{c}{\text{2025}} & \multicolumn{2}{c}{\text{2025}} & \multicolumn{2}{c}{\text{2025}} & \multicolumn{2}{c}{\text{2024}} & \multicolumn{2}{c}{\text{2024}} & \multicolumn{2}{c}{\text{2024}} & \multicolumn{2}{c}{\text{2023}} & \multicolumn{2}{c}{\text{2023}} & \multicolumn{2}{c}{\text{2023}}  \\
 & \textbf{MSE} & \textbf{MAE} & \textbf{MSE} & \textbf{MAE} & \textbf{MSE} & \textbf{MAE} & \textbf{MSE} & \textbf{MAE} & \textbf{MSE} & \textbf{MAE} & \textbf{MSE} & \textbf{MAE} & \textbf{MSE} & \textbf{MAE} & \textbf{MSE} & \textbf{MAE} & \textbf{MSE} & \textbf{MAE}  \\ \midrule
\textbf{ETTh1} & & & & & & & & & & & & & & & & \\ 
96 &\textbf{0.376} &0.401&\underline{0.381} & 0.405 & 0.392 & \textbf{0.390} & 0.389 & 0.402 & 0.386 & 0.405 & 0.386 & \underline{0.395} & 0.414 & 0.419 & 0.423 & 0.448 & 0.384 & 0.402  \\ 
192 &\textbf{0.414}&\textbf{0.421}& \underline{0.430} & 0.434 & 0.449 & 0.439 & 0.435 & 0.440 & 0.441 & 0.436 & 0.437& \underline{0.424} & 0.460& 0.445 & 0.450 & 0.471 & 0.474 & 0.429\\ 
336&\textbf{0.432}&\textbf{0.435} & \underline{0.443} & \underline{0.446} & 0.467 & 0.481 & 0.450 & 0.448 & 0.487 & 0.458 & 0.479 & 0.446 & 0.501 & \underline{0.466} & 0.570 & 0.546 & 0.491 & 0.469  \\ 
720&\textbf{0.463}& \underline{0.473}& 0.502 & 0.501 & \underline{0.475} & 0.468 & 0.480 & \textbf{0.465} & 0.503 & 0.491 & 0.481 & 0.470 & 0.500 & 0.488 & 0.653 & 0.621 & 0.521 & 0.500 \\ \midrule
\textbf{ETTh2} & & & & & & & & & & & & & & & & \\ 
96&\textbf{0.279}& \underline{0.340}& 0.291 & 0.346 & 0.292 & 0.357 & 0.296 & 0.349 & 0.297 & 0.349 & \underline{0.288} & \textbf{0.338} & 0.302 & 0.348 & 0.745 & 0.584 & 0.340 & 0.374 \\ 
192&\textbf{0.362}&\underline{0.395} & \underline{0.369} & 0.397 & 0.380 & 0.402 & 0.371 & 0.400 & 0.380 & 0.400 & 0.374 & \textbf{0.390} & 0.388 & 0.400 & 0.877 & 0.656 & 0.402 & 0.414 \\ 
336&\textbf{0.337}& \textbf{0.385}& \underline{0.374} & \underline{0.414} & 0.391 & 0.420 & 0.402 & 0.449 & 0.428 & 0.432 & 0.415 & 0.426 & 0.426 & 0.433 & 1.043 & 0.731 & 0.452 & 0.452 \\ 
720&\textbf{0.395}&\textbf{0.430} & \underline{0.419} & 0.447 & 0.437 & 0.455 & 0.425 & \underline{0.438} & 0.427 & 0.445 & 0.420 & 0.440 & 0.431 & 0.446 & 1.104 & 0.763 & 0.462 & 0.468  \\ \midrule
\textbf{ETTm1} & & & & & & & & & & & & & & & &  \\ 
96&\textbf{0.289}& \textbf{0.343}& \underline{0.298} & \underline{0.355} & 0.311 & 0.380 & 0.312 & 0.371 & 0.334 & 0.368 & 0.355 & 0.376 & 0.329 & 0.367 & 0.404 & 0.426 & 0.338 & 0.375 \\ 
192&\textbf{0.338}&\textbf{0.371} & \underline{0.347} & \underline{0.381} & 0.389 & 0.419 & 0.365 & 0.409 & 0.377 & 0.391 & 0.391 & 0.392 & 0.367 & 0.385 & 0.450 & 0.451 & 0.374 & 0.387  \\ 
336&\textbf{0.370}&\textbf{0.390} & \underline{0.374} & \underline{0.397} & 0.401 & 0.417 & 0.421 & 0.410 & 0.426 & 0.420 & 0.424 & 0.415 & 0.399 & 0.410 & 0.532 & 0.515 & 0.410 & 0.411 \\ 
720 &\textbf{0.423}&\textbf{0.418}& \underline{0.429} & \underline{0.428} & 0.488 & 0.476 & 0.496 & 0.437 & 0.491 & 0.459 & 0.487 & 0.450 & 0.454 & 0.439 & 0.666 & 0.589 & 0.478 & 0.450  \\ \midrule
\textbf{ETTm2} & & & & & & & & & & & & & & & & \\ 
96&\textbf{0.168}& \underline{0.260}& 0.176 & 0.264 & 0.191 & 0.301 & 0.185 & 0.290 & 0.180 & 0.264 & 0.182 & 0.265 & \underline{0.175} & \textbf{0.259} & 0.287 & 0.366 & 0.187 & 0.267  \\ 
192&\underline{0.235}&\textbf{0.298} & \textbf{0.231} & 0.303 & 0.253 & 0.312 & 0.292 & 0.309 & 0.250 & 0.309 & 0.246 & 0.304 & 0.241 & \underline{0.302} & 0.414 & 0.492 & 0.249 & 0.309 \\ 
336&\textbf{0.274}&\textbf{0.327} & \underline{0.290} & \underline{0.339} & 0.298 & 0.342 & 0.321 & 0.367 & 0.311 & 0.348 & 0.307 & 0.342 & 0.305 & 0.343 & 0.597 & 0.542 & 0.321 & 0.351  \\ 
720&\textbf{0.376}&\textbf{0.393}& \underline{0.388} & \underline{0.398} & 0.409 & 0.407 & 0.401 & 0.400 & 0.412 & 0.407 & 0.407 & 0.398 & 0.402 & 0.400 & 1.730 & 1.042 & 0.408 & 0.403  \\ \midrule
\textbf{Exchange} & & & & & & & & & & & & & & & & \\ 
96&\textbf{0.085} &\underline{0.205} &0.097 &0.222 &\underline{0.086}&0.206 & 0.089&0.208 &0.091 &0.211 & 0.088&0.209 &0.087 &\textbf{0.202} &0.095 &0.218 &0.093 & 0.211\\ 
192&\textbf{0.179} & \textbf{0.303}& 0.191&0.315 &0.182 &0.304 &0.184 &0.309 &0.182 & 0.303&0.188 & 0.311&\underline{0.180} &0.305 &0.193 &0.318 &0.194 &0.315  \\ 
336& \textbf{0.311}&\textbf{0.402} & 0.337&0.424 &0.330&0.416&0.333&0.416 &0.337 & 0.421&0.346 & 0.423& \underline{0.318}& \underline{0.407}&0.359 &0.429 &0.358 &0.433 \\ 
720& \textbf{0.858}&\textbf{0.696}&0.877 & 0.706&0.865 & 0.702& 0.870&\underline{0.701}&\underline{0.862} &0.703 & 0.913& 0.717&0.863 &0.703 &0.918 &0.721 &0.880 &0.719 \\\midrule
\textbf{Weather} & & & & & & & & & & & & & & & &  \\ 
96&\textbf{0.150}& \textbf{0.199}& \underline{0.153} & \underline{0.205} & 0.169 & 0.221 & 0.174 & 0.218 & 0.174 & 0.214 & 0.192 & 0.232 & 0.177 & 0.218 & 0.158 & 0.230 & 0.172 & 0.220 \\ 
192&\textbf{0.194}&\textbf{0.242} & \underline{0.196} & \underline{0.244} & 0.205 & 0.248 & 0.200 & 0.258 & 0.221 & 0.254 & 0.240 & 0.271 & 0.225 & 0.259 & 0.206 & 0.277 & 0.219 & 0.261  \\ 
336&\underline{0.252}&\underline{0.288} & \textbf{0.246} & \textbf{0.283} & 0.288 & 0.299 & 0.280 & 0.299 & 0.278 & 0.296 & 0.292 & 0.307 & 0.278 & 0.297 & 0.272 & 0.335 & 0.280 & 0.306  \\ 
720&\textbf{0.313}&\textbf{0.333} & \underline{0.314} & \underline{0.334} & 0.335 & 0.369 & 0.352 & 0.359 & 0.358 & 0.347 & 0.364 & 0.353 & 0.354 & 0.348 & 0.398 & 0.418 & 0.365 & 0.359 \\ \midrule
\textbf{ECL} & & & & & & & & & & & & & & & &  \\ 
96&\textbf{0.134}& \textbf{0.231}& \underline{0.141} & \underline{0.239} & 0.157 & 0.255 & 0.156 & 0.240 & 0.148 & 0.240 & 0.201 & 0.281 & 0.181 & 0.270 & 0.219 & 0.314 & 0.168 & 0.272 \\ 
192&\textbf{0.153}& \textbf{0.248}& \underline{0.159} & \underline{0.255} & 0.188 & 0.271 & 0.161 & 0.268 & 0.162 & 0.253 & 0.201 & 0.283 & 0.188 & 0.274 & 0.231 & 0.322 & 0.184 & 0.289  \\ 
336&\textbf{0.170}&\textbf{0.266} & \underline{0.171} & \underline{0.268} & 0.192 & 0.275 & 0.195 & 0.272 & 0.178 & 0.269 & 0.215 & 0.298 & 0.204 & 0.293 & 0.246 & 0.337 & 0.198 & 0.300  \\ 
720&\textbf{0.201}&\textbf{0.293} & \underline{0.208} & \underline{0.300} & 0.241 & 0.339 & 0.231 & 0.307 & 0.225 & 0.317 & 0.257 & 0.331 & 0.246 & 0.324 & 0.280 & 0.363 & 0.220 & 0.320 \\ %\midrule
% Average& 0.304&0.349&0.315&
 \bottomrule
% Continue for Weather, ECL, and Traffic
\end{tabular}
}
\caption{Comprehensive comparison across various datasets, prediction horizons, and baselines. The \textbf{bolded} results denote the best performance, and the \underline{underlined} results indicate the second best.}
\label{tab:performance}
\end{table*}
\subsection{Cross-Variate Global Context}
\label{sec:CVGlobal}
%When analyzing multivariate time-series data, the human brain naturally identifies and compares global patterns across variables first, storing this information to help focus on local details later. 
%\textcolor{red}{Need a complete overhaul}
The global patches $\mathbf{\tilde{X}}_g$ is first passed through the global feature extractor, which is a dual Mamba system, responsible for cross-variate global feature extraction. %This inspires our approach, which
We begin by concatenating along the first and second dimension of $\mathbf{\tilde{X}}_g$, viewed with a new shape $\mathbf{\tilde{X}}_g\in \mathbb{R}^{(D* PN_g)\times PL_g}$ as illustrated in Figure \ref{fig:patch}. This allows the learning of variate-level correlation across all $D$ dimensions as the selection mechanism of Mamba will filter the relevant variates and patches, enabling the global model to capture cross-variate global context. However, Mamba processes data unilaterally, attending only to antecedent patches, which limits learning of the full global context. Inspired by S-Mamba \citep{wang2025mamba}, we employ two Mamba models to scan the sequence in both forward and backward directions before aggregating the results. This approach improves the learning of correlations between global patches across variables. Specifically, we have 
\begin{align}
    \overrightarrow {\mathbf{Z}_g} &= \overrightarrow{\text{Mamba Layers}} (\mathbf{\tilde{X}}_g),\\
    \overleftarrow {\mathbf{Z}_g} &= \overleftarrow{\text{Mamba Layers}} (\overleftarrow{\mathbf{\tilde{X}}}_g),\\
    \mathbf{Z}_g &= \overrightarrow {\mathbf{Z}_g}+\overleftarrow {\mathbf{Z}_g},
\end{align}
where $\overleftarrow{\mathbf{\tilde{X}}}_g \in \mathbb{R}^{(D*PN_g)\times PL_g}$ is obtained by reversing the the first dimension of $\mathbf{\tilde{X}}_g$. The output of the global model  $\mathbf{Z}_g\in \mathbb{R}^{(D*PN_g)\times d_\text{model}}$, which serves as an intermediary output of the entire architecture, is then fed to the local model. This intermediary output encapsulates both cross-variate and global context information. %patches from all variables to be combined into a single sequence, as illustrated in the figure. In order to facilitate communication across variables, we have prolonged the sequence length of the patched time series by factor of $D$. 

%However, the computational load of global attention escalates exponentially with the increased sequence length, rendering transformer-based models impractical for large dimensional data. On the other hand, the selective mechanism of Mamba can discern the significance of each patch akin to attention (cite) but with a computational overhead only escalating in a near-linear fashion. Therefore, employing Mamba as the global model and scanning the sequence of patches for all variables in one sweep becomes efficient. 

Note that $d_\text{model}$ represents the model dimension of Mamba, which aligns with the model dimension of the local model discussed in the next section.

\begin{figure*}[t]
    \centering
    % Subplot 1
    \begin{subfigure}[t]{0.24\textwidth} % 24% of the width
        \centering
        \includegraphics[width=\textwidth]{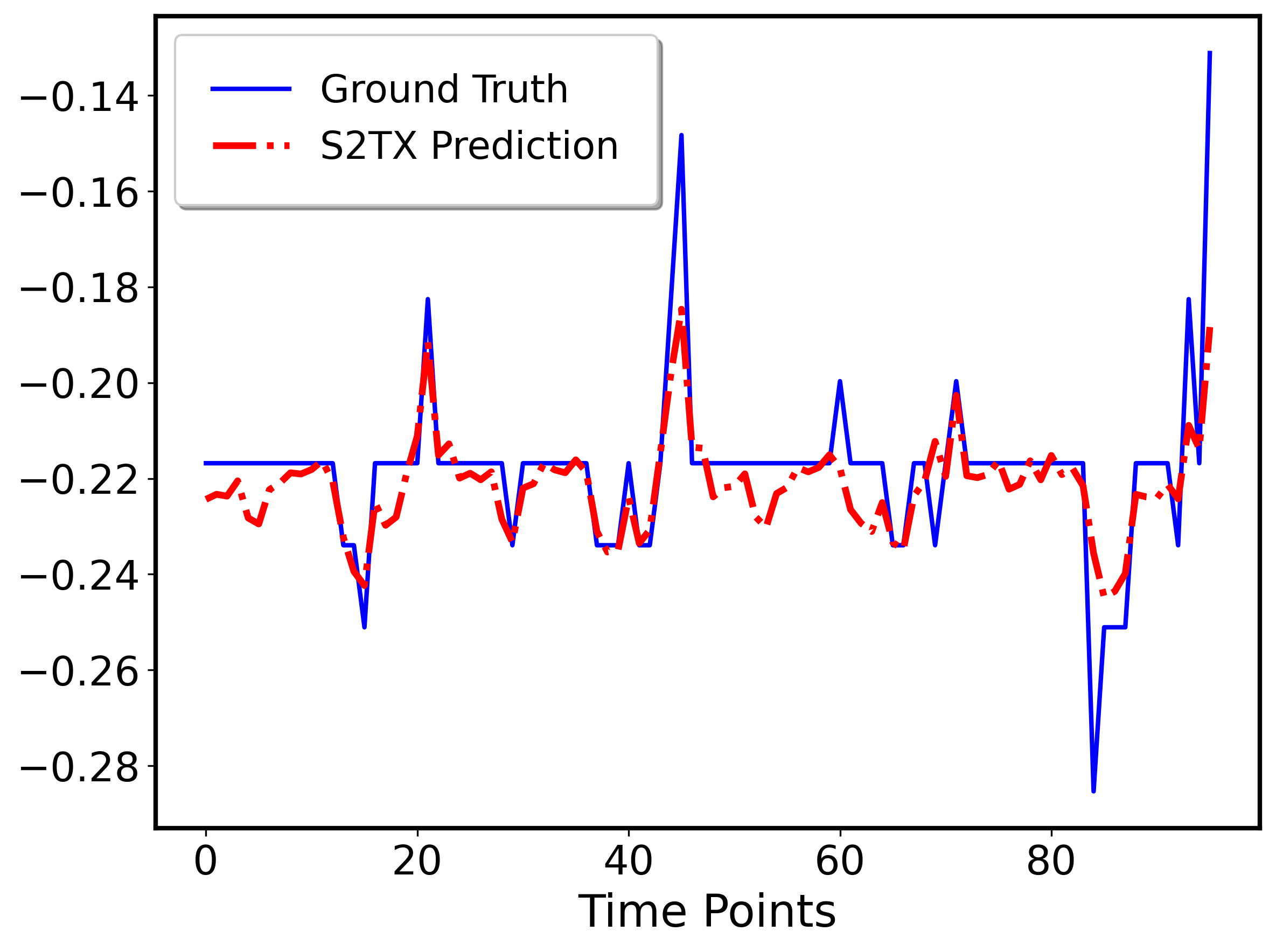} % Replace with your image
        % \caption{Sine Function}
        \label{fig:sine}
    \end{subfigure}
    % Subplot 2
    \begin{subfigure}[t]{0.24\textwidth}
        \centering
        \includegraphics[width=\textwidth]{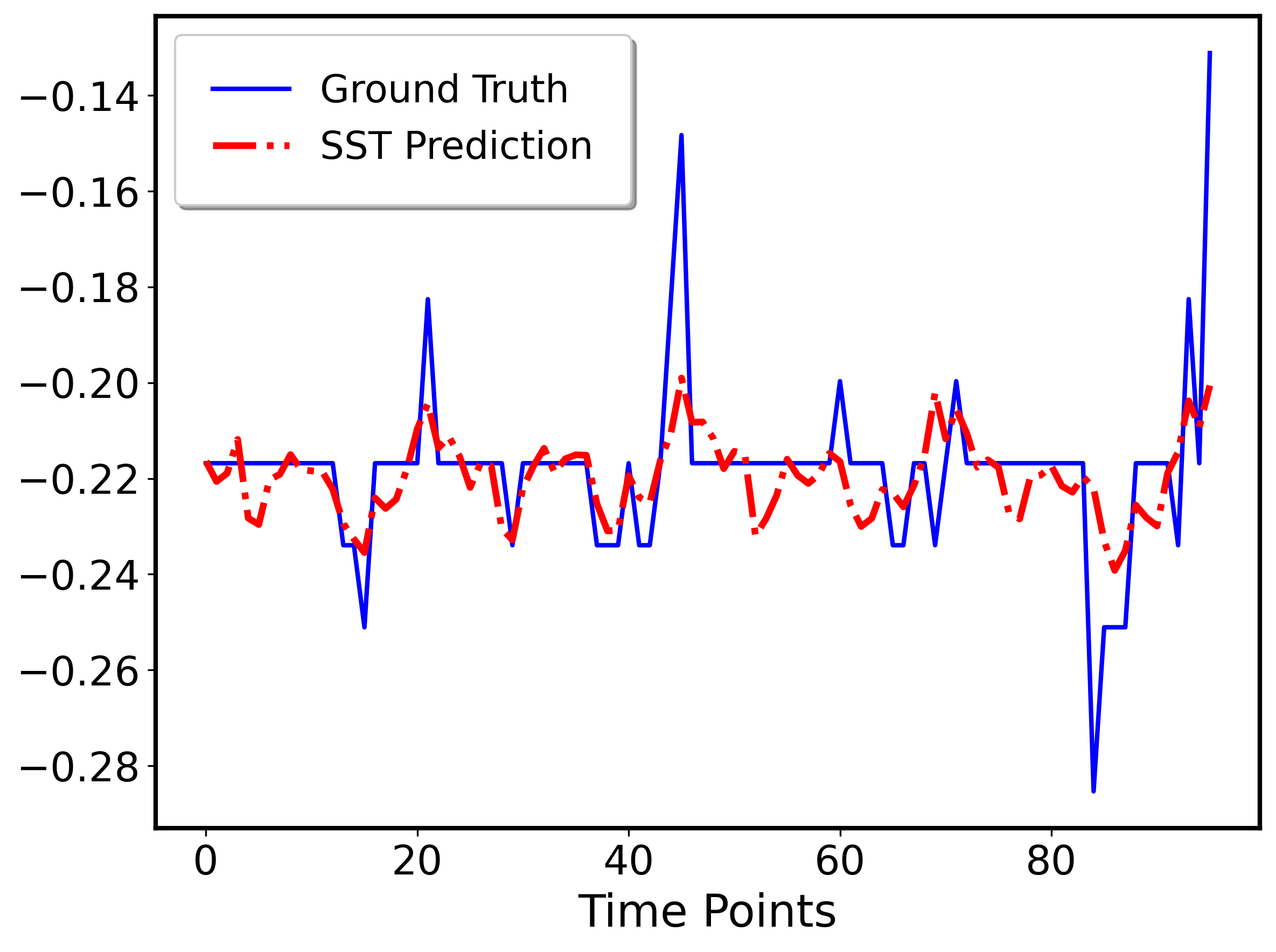} % Replace with your image
        % \caption{Cosine Function}
        \label{fig:cosine}
    \end{subfigure}
    % Subplot 3
    \begin{subfigure}[t]{0.24\textwidth}
        \centering
        \includegraphics[width=\textwidth]{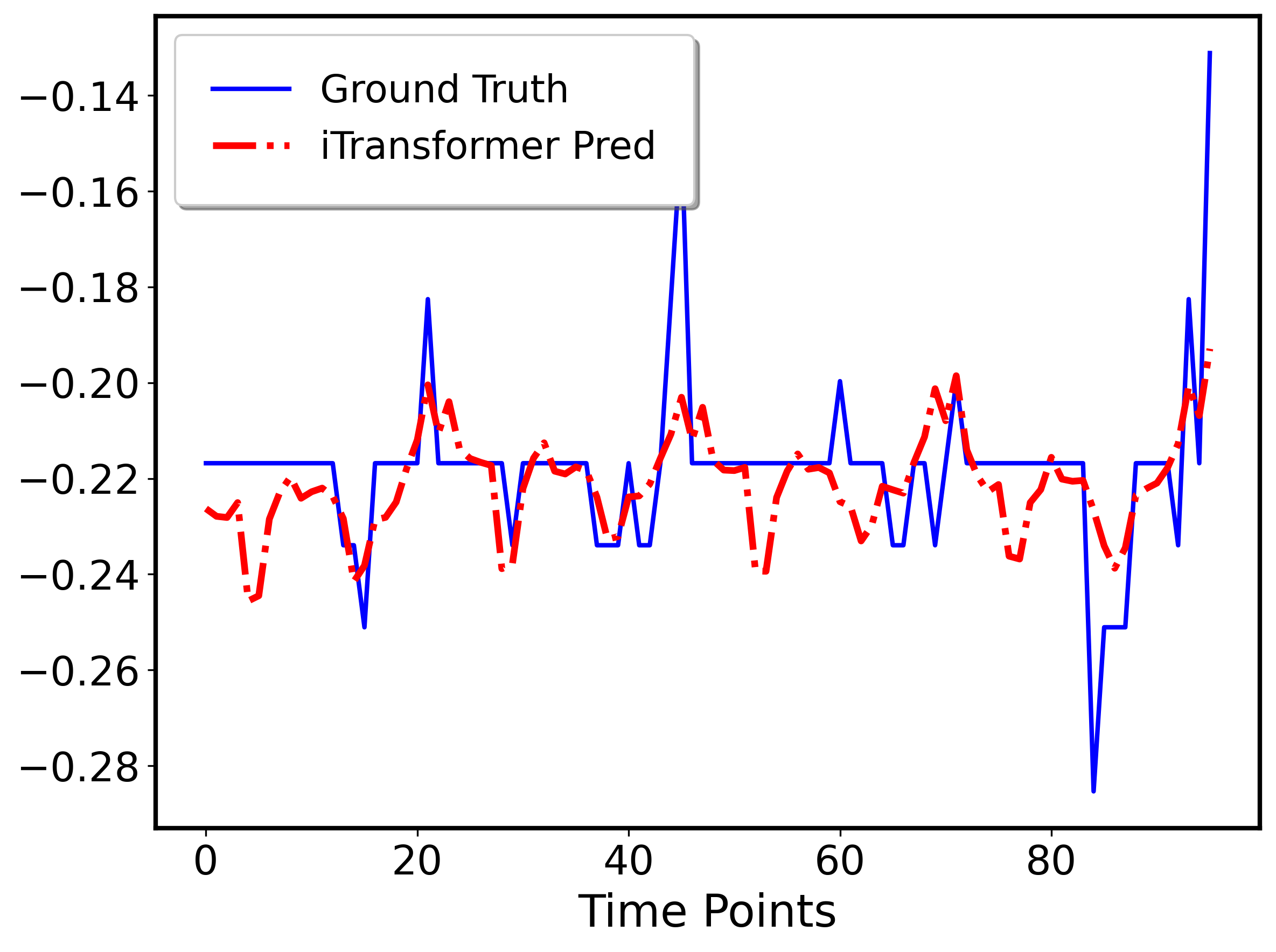} % Replace with your image
        % \caption{Tangent Function}
        \label{fig:tangent}
    \end{subfigure}
    % Subplot 4
    \begin{subfigure}[t]{0.24\textwidth}
        \centering
        \includegraphics[width=\textwidth]{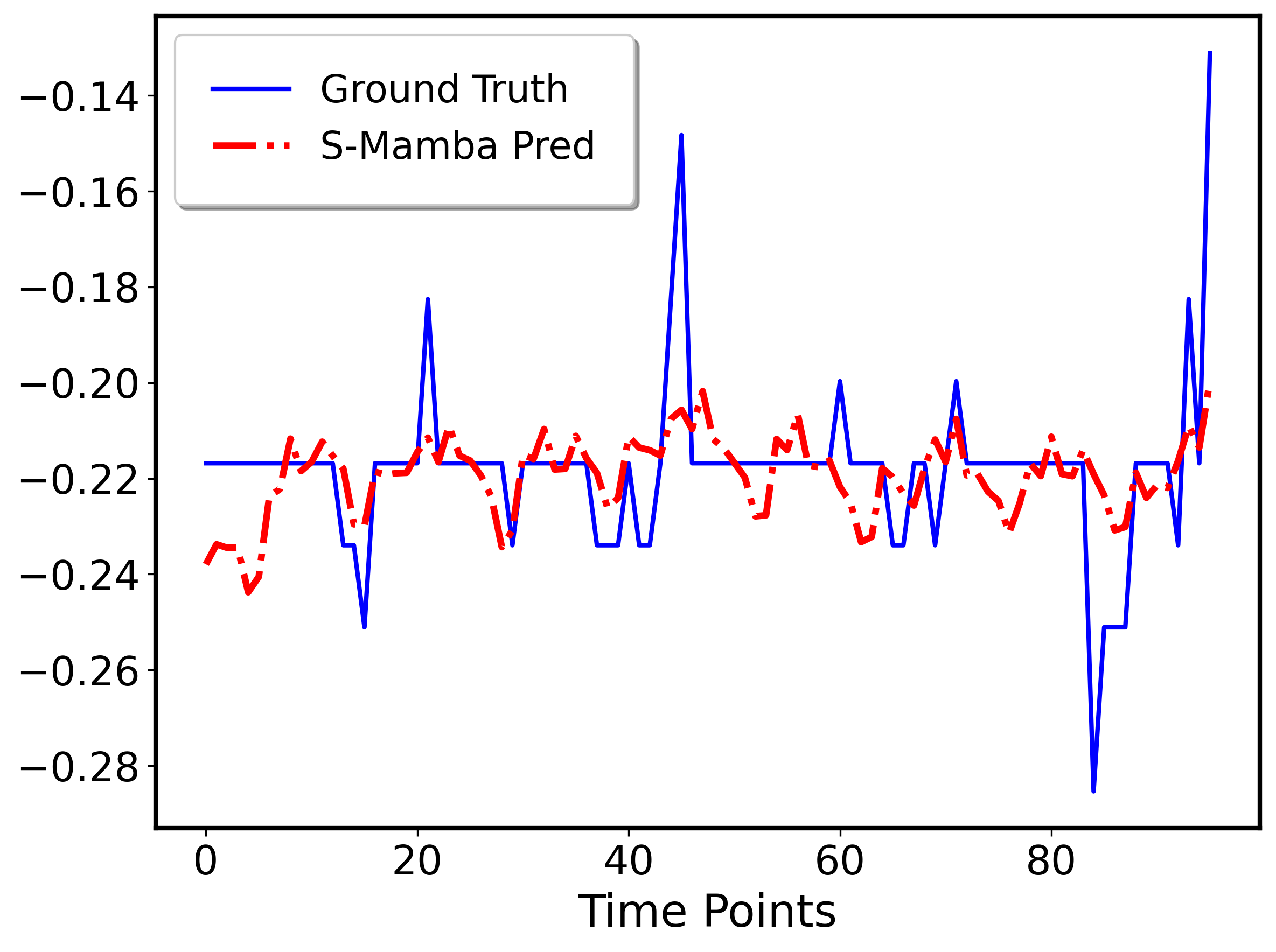} % Replace with your image
        % \caption{Exponential Decay}
        \label{fig:exponential}
    \end{subfigure}
    \caption{Empirical time series versus predicted time series across different architecture. S2TX can better capture the variation of the variable over time. }
    \label{fig:predictTScompare}
\end{figure*}
\subsection{Cross-Attention Local Context}
\label{sec:CALocal}
%\HM{Notation in this subsection is too cluttered, need to somehow simplify.}
With global and cross-variate patterns as context information, the local model can more effectively capture local features and interpret local variations. To this end, we employ a decoder-like transformer with each layer composed of a self-attention without causal masking followed by a cross-attention. Since cross-variate correlation is already captured by the context features, we now take each variate (in the first dimension) of $\mathbf{\tilde{X}}_l\in \mathbb{R}^{D\times PN_l\times PL_l}$ individually as the input of the self-attention to relieve the computation burden of transformer. Denote the $d$-th variate of $\mathbf{\tilde{X}}_l$ after linear projection to $d_{\text{model}}$-dimension by $\mathbf{\tilde{X}}_l^d\in \mathbb{R}^{PN_l\times d_{\text{model}}}$. Similarly, the context feature $\mathbf{Z}_g\in \mathbb{R}^{(D*PN_g)\times d_\text{model}}$ is viewed back to $\mathbf{Z}_g\in \mathbb{R}^{D\times PN_g\times d_\text{model}}$ and the $d$-th variate $\mathbf{Z}_g^d\in \mathbb{R}^{PN_g\times d_\text{model}}$ is sent to the cross-attention as key and value to match the dimension of $\mathbf{\tilde{X}}_l^d$. Specifically, the cross-attention mechanism operates as follows:
\begin{align}
    &\text{AttentionBlock}(\mathbf{\tilde{X}}_l^d, \mathbf{Z}_g^d) \nonumber \\
    &= \text{CrossAttention}(\mathbf{Z}_g^d, \;\text{SelfAttention}(\mathbf{\tilde{X}}_l^d)),
\end{align}
Note that we have omitted the skip connection and normalization steps for a concise presentation. The rest of the local model is the same as a regular transformer decoder as shown in the figure. 

Denote the output of the local model of the $d$th variable to be $\mathbf{Y}_\text{out}^d\in \mathbb{R}^{PN_l\times d_{\text{model}}}$. Stacking the outputs of all variables, we obtain $\mathbf{Y}_\text{out}\in \mathbb{R}^{D\times PN_l\times d_\text{model}}$. The last two dimensions of $\mathbf{Y}_\text{out}$ are then flattened and a final linear head is employed to project from dimension $PN_l\times d_\text{model}$ to $H$, which is the target horizon window. 

\subsection{Runtime Complexity Analysis}
\label{sec:runtime}
% The runtime complexity is primarily influenced by the patch number $PN$, rather than the look-back window length $L$. The Mamba layers, which exhibit linear complexity, process a sequence of length $D \cdot PN$ in a runtime complexity of $O(D \cdot PN)$. In contrast, the Transformer has quadratic complexity and processes $D$ sequences of length $PN$ each, leading to a runtime complexity of $O(D \cdot PN^2)$. Consequently, the overall complexity of the model is $O(D \cdot PN^2)$ with respect to $PN$. To manage complexity as the look-back window length increases, it is reasonable to proportionally increase both the patch length and stride such that $PN$ remains constant. So, when the stride is chosen to be proportional to the length $L$, the overall complexity of S2TX is constant.
The Mamba layers, which exhibit linear complexity, process a sequence of length $D \cdot PN_g$ with a complexity of $O(D \cdot PN_g)$, which is linear to input time series length $L$ due to definition of $PN$ in Equation~\eqref{eq:pn}. On the other hand, while transformer models exhibit quadratic complexity with respect to sequence length, S2TX uses a local look-back window with fixed length $S$, resulting in a complexity of $O(D\cdot PN_l^2) = O(D)$ as $PN_l=O(S)=O(1)$.  Thus, S2TX has an overall linear complexity with respect to $L$ and $D$. Moreover, as $L$ increases, we can proportionally increase both the patch length and stride so that $PN_g$ remains constant, in which case, the overall complexity of S2TX reduces to constant order with respect to $L$ while remaining linear order with respect to $D$. Our empirical results in Section~\ref{sec:runtime} verifies this, showing that S2TX's runtime barely increases with $L$.
% To manage this complexity as the look-back window length $L$ increases, it is reasonable to proportionally increase both the patch length and stride so that $PN$ remains constant, in which case, the overall complexity of S2TX remains constant. Otherwise, if the stride remains fixed for a prolonged look-back window, the overall complexity increases linearly with $L$.

\section{Experiment}
\label{sec:experiment}
We empirically demonstrate that utilizing cross-variate correlation and global-local interaction can significantly improve the forecasting performance. We first introduce the experimental setup, then we showcase the performance of S2TX over a variety of benchmark against recent state-of-the-art architectures. We then demonstrate the efficacy of the main component of S2TX with a set of ablation study and a robustness study where we test the robustness of S2TX with sequences of missing values. Finally, we showcase the low memory footprint and efficient runtime of S2TX compared to Transformer and Mamba in general.

\textbf{Dataset.} 
We benchmark our proposed algorithm S2TX on a set of 7 real-world multivariate time series datasets, including the four Electricity Transformer Temperature datasets ETTh1, ETTh2, ETTm1, and ETTm2, Weather, Electricity, and Exchange rate datasets. Detailed dataset descriptions are provided in the Appendix \ref{appendix:data_desc}.

\textbf{Baselines.} 
We benchmark our proposed algorithm S2TX against most competitive time series forecasting models within three years, including MOE-based model SST \citep{xusst}, Mamba-based models S-Mamba \citep{wang2025mamba} and TimeMachine (TimeM) \citep{ahamed2024timemachine}, transformer-based models iTransformer (iTrans) \citep{liu2023itransformer}, PatchTST \citep{nie2022time}, Crossformer (CrossF) \citep{zhang2023crossformer}, and FEDformer \citep{zhou2022fedformer}, linear-based models RLinear \citep{li2023revisiting} and DLinear \citep{zeng2023transformers}, and TCN-based model TimesNet \citep{wu2022timesnet}. Due to space constraints, the comparisons against DLinear and FEDformer (pre-2023 models) are presented in Appendix \ref{appendix:full_comparison} 

\textbf{Experimental Setting and Metrics.}
For a fair comparison, the experimental setting of all baselines follows the experiment setup of the current SOTA SST. In addition, we use the same hyperparameters as in SST, including global and local patch length, stride, and look-back window. Specifically, we set $PL_g = 48$, $STR_g = 16$, $PL_l = 16$, $STR_l = 8$, and $L = 2S = 336$. For Exchange rate dataset, we use a smaller patch length, stride, and look-back window: $PL_g = 16$, $STR_g = 8$, $PL_l = 4$, $STR_l = 2$, $L = 2S = 192$. The forecast horizon is set to $\{96,192,336,720\}$ for each dataset. We use mean squared error and mean absolute error as metrics to compare performances of different architectures. 

We now present the numerical result of our comprehensive experiments, as well as an ablation study to showcase the importance of each module, and a computation efficiency study comparing canonical architectures, SST, and S2TX. 

\subsection{Benchmark Results}
\label{sec:result}
The performance of 9 different architectures on 7 benchmark datasets and 4 different prediction horizons is presented in Table \ref{tab:performance}. \textbf{Our method S2TX achieves SOTA performance across all benchmark datasets}. In particular, compared to the previous SOTA model SST, S2TX demonstrates consistent improvements on most datasets and performs on par on the weather dataset. For instance, S2TX achieves an $8.4\%$ improvement on the ETTh1 dataset with a prediction horizon of $720$. Moreover, even on the weather dataset, S2TX significantly surpasses other baseline models. The SOTA performance, together with our ablation studies in section \ref{sec:ablation}, suggests that the two novel aspect of S2TX, the cross-variate global features, and the cross-attentional local features, are indeed important for accurately forecasting multivariate time series. 

Guided by cross-variate global context, S2TX demonstrates a superior ability to capture local variations. Figure \ref{fig:predictTScompare} presents a random segment of test time prediction from the electricity dataset on a randomly selected variate, comparing the performance of S2TX, SST, iTransformer, and S-Mamba. S2TX precisely approximates abrupt spikes while the accurate predictions of local variation are less apparent in predictions of other models.
%\paragraph{Qualitative Results}

%To intuitively illustrate the improvements of S2TX, we present a visual graph showcasing a random prediction segment from the electricity dataset on a randomly selected variable, comparing the performance of S2TX, SST, iTransformer, and iMamba. Guided by cross-dimensional global context, S2TX demonstrates a superior ability to capture local variations, as evidenced by its precise approximation of abrupt spikes—an ability less apparent in the other models.

\subsection{Ablation and Robustness Studies}
\label{sec:ablation}
\begin{figure}[t]
    \centering
    \includegraphics[width=0.45\textwidth]{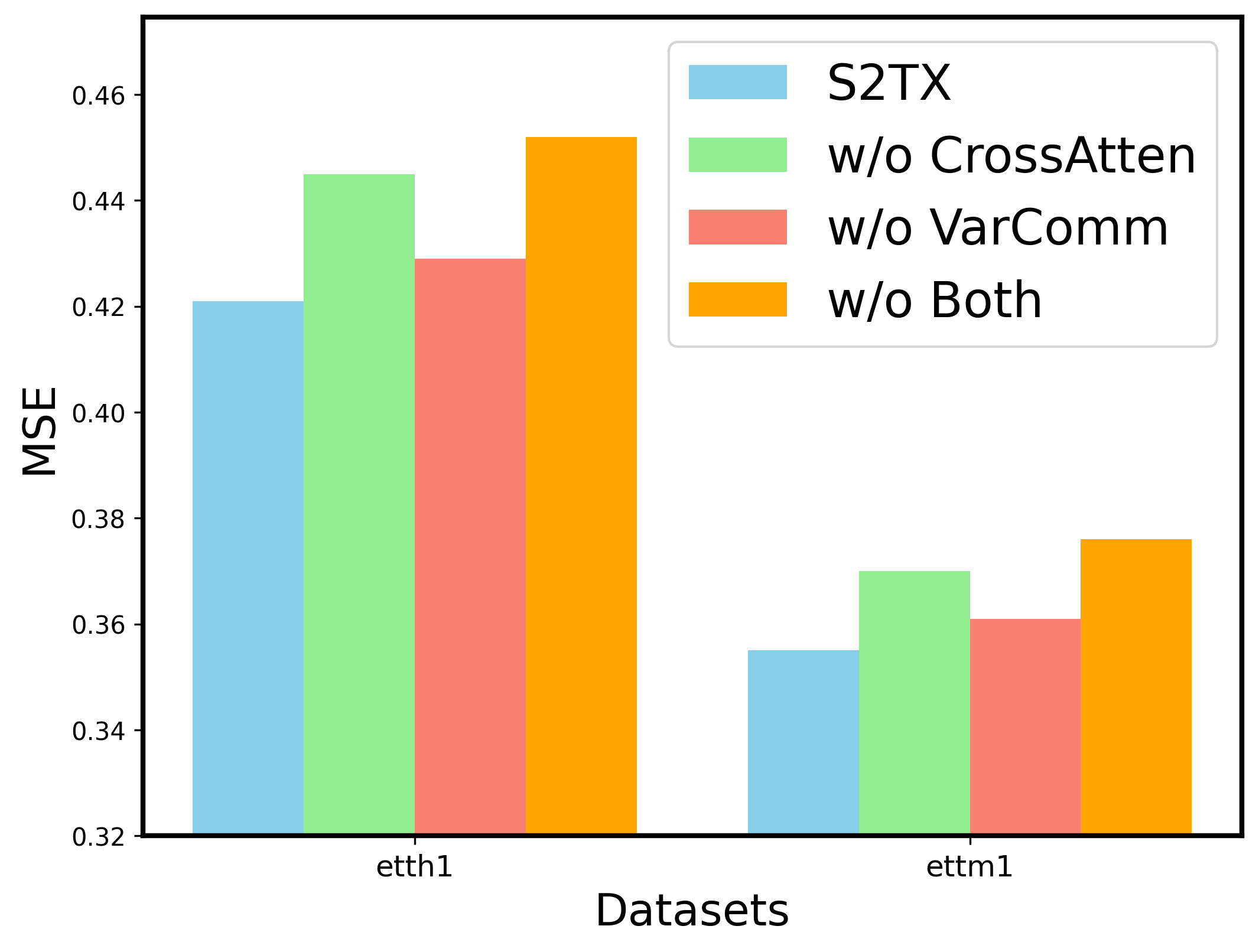} % Change to your image file
    \caption{Ablation study on different components of S2TX tested on ETTh1 and ETTm1 datasets. The efficacy of each component of the proposed architectures is measured by the degradation of performance after each (or both) component(s) was excluded. }
    \label{fig:ablation}
\end{figure}
\textbf{Ablation on Model Components.}
We perform ablation studies by removing key components of S2TX. To first assess the impact of cross-variate communication in learning global context, we input the patch sequence of each variate separately into the global model, rather than using the concatenated cross-variate patch sequence. Second, to evaluate the effectiveness of the context-local cross-attention mechanism, we remove cross-attention and instead concatenate the global context and local features before the final linear head. Finally, we remove both mechanisms to evaluate their combined effect. We conduct ablation studies on the ETTh1 and ETTm1 datasets and report the MSE metric, averaged across four different prediction lengths. As shown in Figure \ref{fig:ablation}, the global-local cross-attention contributes the most to the overall improvement of S2TX, while variable communication also positively influences the results.

\textbf{Robustness to Missing Values.}
In real-world multivariate time series datasets, it is common to observe missing values. Unlike traditional tabular data where a few elements are missing, missing values in time series could exist for small periods of sequences. In this set of robustness experiments, we randomly select small sequences of 4 time steps to be missing and interpolate these randomly missing periods with the value of the last observed time step. In Table \ref{tab:robustness}, we present the MSE of different architectures under various percentage of missing values. We show that S2TX, with the addition of cross-variate global context and the cross-attentional global-local feature interplay, is highly robust compared to SST, which showed much-worsened degradation as the percentage of missing value increases. 
\begin{table}[t]
    \centering
    \footnotesize
    \renewcommand{\arraystretch}{1.1}
    \setlength{\tabcolsep}{8pt}
    \begin{tabular}{lcc}  % l = left, c = center, r = right
        \toprule
        \textbf{Miss Ratio} & \textbf{S2TX}& \textbf{SST}\\
        \midrule
        0\%         &0.421(-0.0\%)  &0.439(-0.0\%)\\
        4\%         &0.424(-0.7\%)  & 0.440(-0.2\%) \\
        8\%         & 0.425(-0.9\%) & 0.443(-0.9\%) \\
        16\%        & 0.424(-0.7\%) & 0.450(-2.5\%) \\
        24\%        & 0.429(-1.9\%) &0.468(-6.6\%) \\
        32\%        &0.431(-2.3\%)  & 0.471(-7.0\%)\\
        40\%        & 0.441(-4.7\%) & 0.499(-13.4\%)\\
        \bottomrule
    \end{tabular}
    \caption{Performance on ETTh1 with increasing proportion of missing values; results are MSE averaged over all four prediction horizons.}
    \label{tab:robustness}
\end{table}

\subsection{Memory and Runtime Analysis}\label{sec:runtime}
To ensure a fair runtime comparison, we evaluate S2TX alongside SST, the vanilla Transformer, and Mamba on a single NVIDIA RTX 6000 Ada Generation GPU. The two versions of S2TX and SST correspond to configurations with either a fixed patch number or a fixed stride length. In the former case, the patch number remains constant regardless of sequence length, whereas in the latter case, the patch number increases proportionally with sequence length.

It is important to note that we compare against the vanilla Transformer and Mamba, rather than their inverted versions, as the respective attention and selective mechanisms in the inverted versions operate on the variate dimension. As the look-back window sequence length increases, the memory usage and runtime of the Transformer grow exponentially, reaching the GPU's memory limit when the sequence length hits 2000. In contrast, Mamba scales linearly in both memory and time metrics.

\begin{figure}[t]
    \centering
    % Subfigure 1
    \begin{subfigure}[]{0.6\columnwidth} % 45% of the width for the first subfigure
        \centering
        \includegraphics[width=\linewidth]{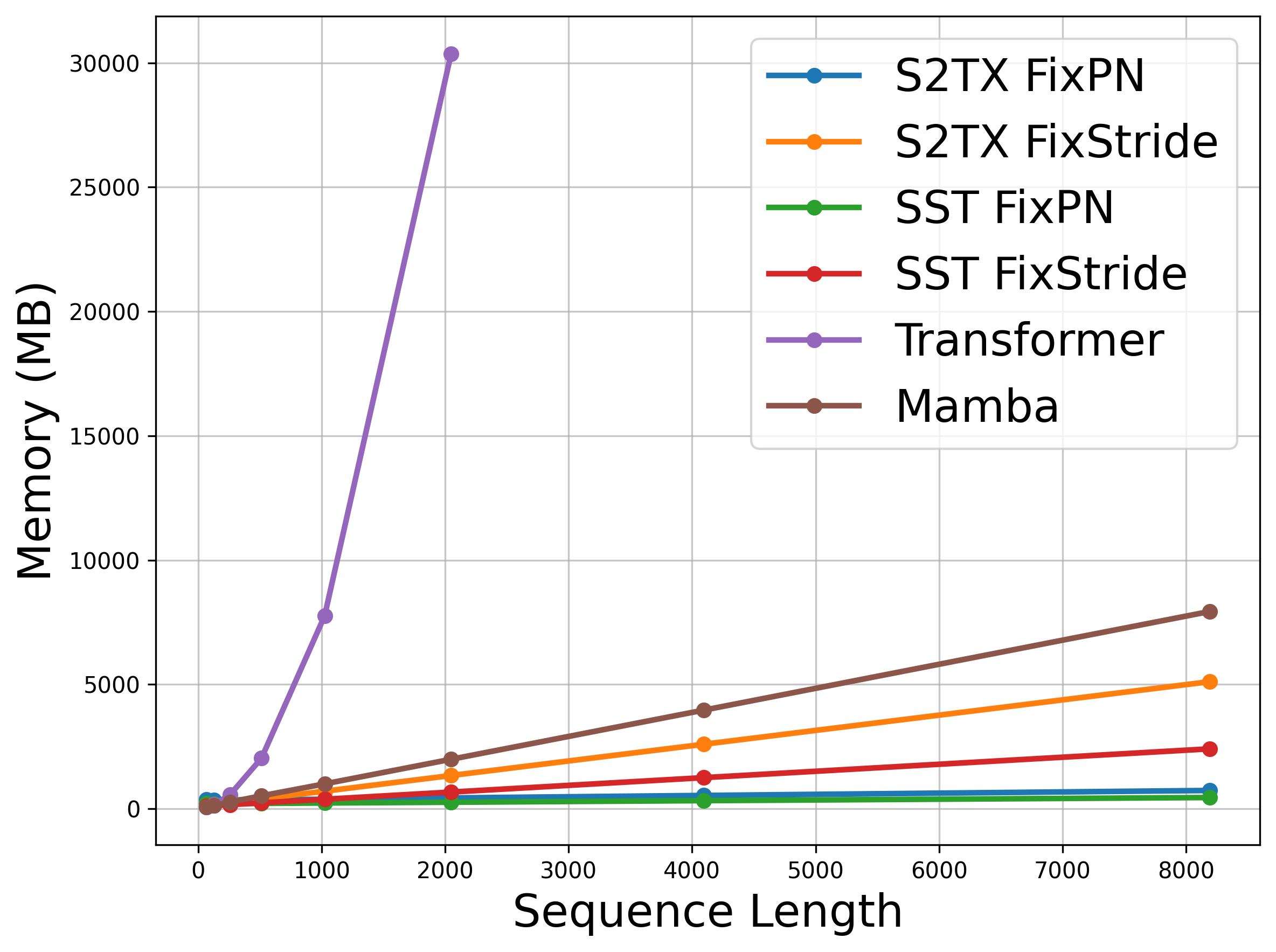} % Replace with your image file
        \caption{}%Memory usage comparison between S2TX, SST, vanilla transformer and Mamba.}
        \label{fig:subfigure1}
    \end{subfigure}
    \quad % Adds horizontal spacing between the two subfigures
    % Subfigure 2
    \begin{subfigure}[]{0.6\columnwidth} % 45% of the width for the second subfigure
        \centering
        \includegraphics[width=\linewidth]{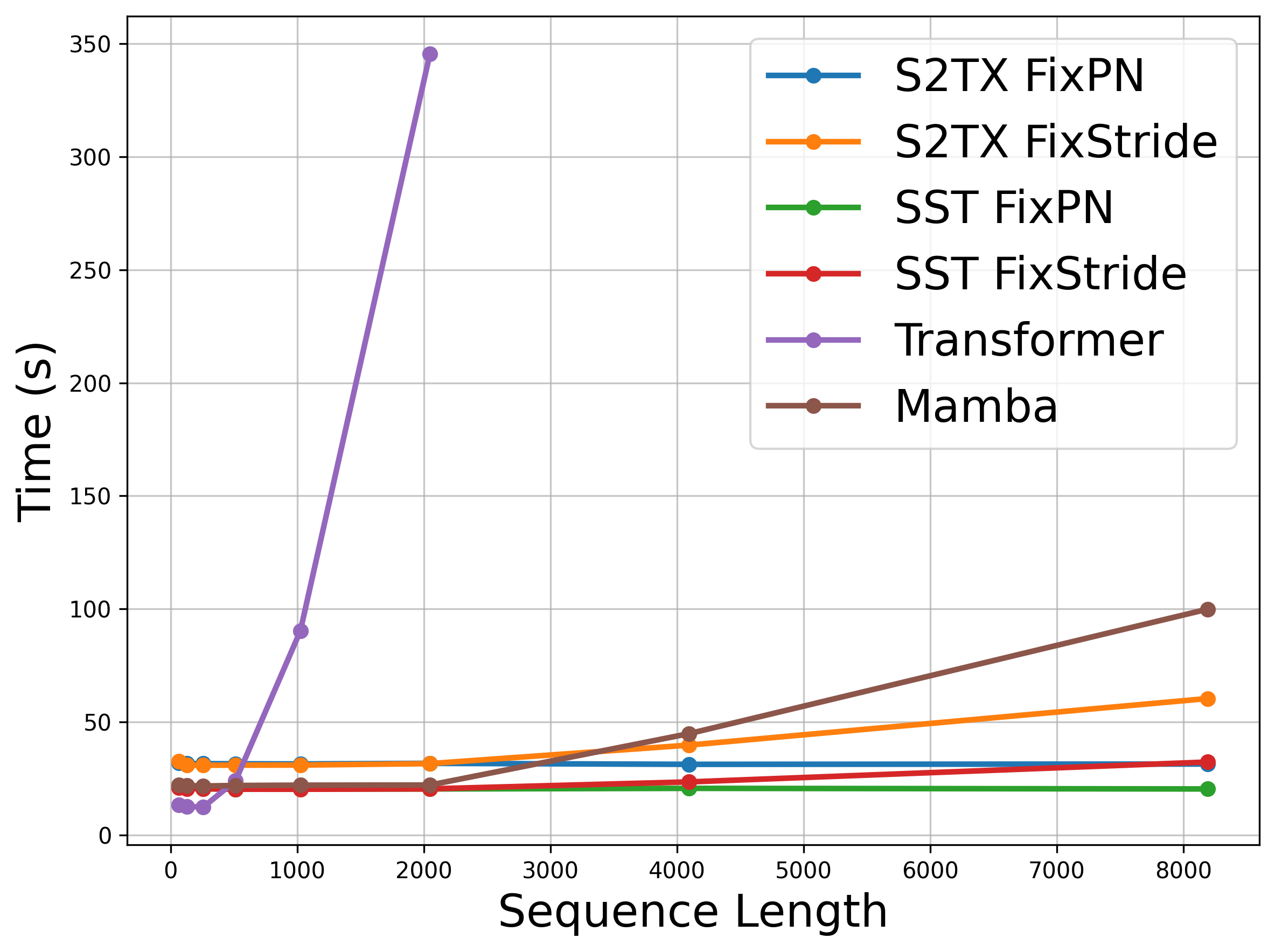} % Replace with your image file
        \caption{}%Time complexity comparison between S2TX, SST, vanilla transformer and Mamba.}
        \label{fig:subfigure2}
    \end{subfigure}
    % Main caption for the figure
    \caption{Memory and run-time comparison between S2TX and other canonical architectures.}
    \label{fig:ablation_study}
\end{figure}

Both S2TX and SST scale more efficiently than Mamba, owing to the fixed short local look-back window combined with the patching technique, which effectively reduces the sequence length by a factor of the stride length. The complexity experiment result is presented in Figure \ref{fig:ablation_study}. Consistent to the runtime analysis in section~\ref{sec:runtime}, when the global patch number is fixed, both S2TX and SST achieve nearly constant runtime complexity. However, when comparing S2TX to SST, SST scales slightly better due to the additional cross-attention mechanism in S2TX.

\section{Discussion and Future Work}
In this work, we introduce a new architecture, \emph{State-Space Transformer with cross-attention} (S2TX), for multivariate time series modeling. We first noted that the multi-scale patching methods, although enhance the learning of temporal dependencies, neglect the cross-variate correlation--a crucial aspect of multivariate time series modeling. Also, global and local patches are processed independently, overlooking the global and local interactions that occur in many real-world scenarios. We propose a novel cross-attention based architecture that integrates state space models and transformers. This cross-attention architecture, combined with patchification, fully leverages the strengths of Mamba and transformers by integrating cross-variate global features from Mamba with the local features of the transformer. Our architecture generally improves over current state-of-the-art in various datasets and 4 different prediction horizons. The SOTA performance of S2TX is not only achieved with a low memory footprint and fast computation runtime but also demonstrated robust performance when facing time series with sequences of missing values. Given these advantages, S2TX unlocks new possibilities for time series forecasting by effectively capturing cross-variate correlations and global-local feature interactions.

\textbf{Limitations.}
Several limitations exist in our current architectures. One key limitation is that cross-variate correlations are not explicitly explored at a local level. Although S2TX maintains low memory usage and fast runtime, incorporating local cross-variate correlations could further enhance performance. Another limitation is the lack of diversity in the multi-scale approach. The current architecture only deals with global and local patches with no learning of the intermediates scales. Intermediate time scales, however, could be important for extremely long sequences where the difference in time scales between global and local contexts is dramatic. Incorporating multiple time scales within an architecture while remaining lightweight is still unsolved. We leave these for future works. 

\paragraph{Impact Statement}
This paper presents work whose goal is to advance the field of Machine Learning. There are many potential societal consequences of our work, none of which we feel must be specifically highlighted here.

% \bibliography{main}

\begin{thebibliography}{34}
\providecommand{\natexlab}[1]{#1}
\providecommand{\url}[1]{\texttt{#1}}
\expandafter\ifx\csname urlstyle\endcsname\relax
  \providecommand{\doi}[1]{doi: #1}\else
  \providecommand{\doi}{doi: \begingroup \urlstyle{rm}\Url}\fi

\bibitem[Ahamed \& Cheng(2024)Ahamed and Cheng]{ahamed2024timemachine}
Ahamed, M.~A. and Cheng, Q.
\newblock Timemachine: A time series is worth 4 mambas for long-term forecasting.
\newblock \emph{arXiv preprint arXiv:2403.09898}, 2024.

\bibitem[Angryk et~al.(2020)Angryk, Martens, Aydin, Kempton, Mahajan, Basodi, Ahmadzadeh, Cai, Filali~Boubrahimi, Hamdi, et~al.]{angryk2020multivariate}
Angryk, R.~A., Martens, P.~C., Aydin, B., Kempton, D., Mahajan, S.~S., Basodi, S., Ahmadzadeh, A., Cai, X., Filali~Boubrahimi, S., Hamdi, S.~M., et~al.
\newblock Multivariate time series dataset for space weather data analytics.
\newblock \emph{Scientific data}, 7\penalty0 (1):\penalty0 227, 2020.

\bibitem[Bahdanau(2014)]{bahdanau2014neural}
Bahdanau, D.
\newblock Neural machine translation by jointly learning to align and translate.
\newblock \emph{arXiv preprint arXiv:1409.0473}, 2014.

\bibitem[Bloomfield(2004)]{bloomfield2004fourier}
Bloomfield, P.
\newblock \emph{Fourier analysis of time series: an introduction}.
\newblock John Wiley \& Sons, 2004.

\bibitem[Box et~al.(2015)Box, Jenkins, Reinsel, and Ljung]{box2015time}
Box, G.~E., Jenkins, G.~M., Reinsel, G.~C., and Ljung, G.~M.
\newblock \emph{Time series analysis: forecasting and control}.
\newblock John Wiley \& Sons, 2015.

\bibitem[Chen et~al.(2021)Chen, Fan, and Panda]{chen2021crossvit}
Chen, C.-F.~R., Fan, Q., and Panda, R.
\newblock Crossvit: Cross-attention multi-scale vision transformer for image classification.
\newblock In \emph{Proceedings of the IEEE/CVF international conference on computer vision}, pp.\  357--366, 2021.

\bibitem[Durbin \& Koopman(2012)Durbin and Koopman]{durbin2012time}
Durbin, J. and Koopman, S.~J.
\newblock \emph{Time series analysis by state space methods}, volume~38.
\newblock OUP Oxford, 2012.

\bibitem[Gers et~al.(2000)Gers, Schmidhuber, and Cummins]{gers2000learning}
Gers, F.~A., Schmidhuber, J., and Cummins, F.
\newblock Learning to forget: Continual prediction with lstm.
\newblock \emph{Neural computation}, 12\penalty0 (10):\penalty0 2451--2471, 2000.

\bibitem[Gong et~al.(2023)Gong, Tang, and Liang]{gong2023patchmixer}
Gong, Z., Tang, Y., and Liang, J.
\newblock Patchmixer: A patch-mixing architecture for long-term time series forecasting.
\newblock \emph{arXiv preprint arXiv:2310.00655}, 2023.

\bibitem[Graves et~al.(2013)Graves, Mohamed, and Hinton]{graves2013speech}
Graves, A., Mohamed, A.-r., and Hinton, G.
\newblock Speech recognition with deep recurrent neural networks.
\newblock In \emph{2013 IEEE international conference on acoustics, speech and signal processing}, pp.\  6645--6649. Ieee, 2013.

\bibitem[Gu \& Dao(2023)Gu and Dao]{gu2023mamba}
Gu, A. and Dao, T.
\newblock Mamba: Linear-time sequence modeling with selective state spaces.
\newblock \emph{arXiv preprint arXiv:2312.00752}, 2023.

\bibitem[Gu et~al.(2020)Gu, Dao, Ermon, Rudra, and R{\'e}]{gu2020hippo}
Gu, A., Dao, T., Ermon, S., Rudra, A., and R{\'e}, C.
\newblock Hippo: Recurrent memory with optimal polynomial projections.
\newblock \emph{Advances in neural information processing systems}, 33:\penalty0 1474--1487, 2020.

\bibitem[Gu et~al.(2021)Gu, Goel, and R{\'e}]{gu2021efficiently}
Gu, A., Goel, K., and R{\'e}, C.
\newblock Efficiently modeling long sequences with structured state spaces.
\newblock \emph{arXiv preprint arXiv:2111.00396}, 2021.

\bibitem[Horn \& Johnson(2012)Horn and Johnson]{horn2012matrix}
Horn, R.~A. and Johnson, C.~R.
\newblock \emph{Matrix analysis}.
\newblock Cambridge university press, 2012.

\bibitem[Hsieh(2004)]{hsieh2004nonlinear}
Hsieh, W.~W.
\newblock Nonlinear multivariate and time series analysis by neural network methods.
\newblock \emph{Reviews of Geophysics}, 42\penalty0 (1), 2004.

\bibitem[Hyndman(2018)]{hyndman2018forecasting}
Hyndman, R.
\newblock \emph{Forecasting: principles and practice}.
\newblock OTexts, 2018.

\bibitem[Koop et~al.(2010)Koop, Korobilis, et~al.]{koop2010bayesian}
Koop, G., Korobilis, D., et~al.
\newblock Bayesian multivariate time series methods for empirical macroeconomics.
\newblock \emph{Foundations and Trends{\textregistered} in Econometrics}, 3\penalty0 (4):\penalty0 267--358, 2010.

\bibitem[Lai et~al.(2018)Lai, Chang, Yang, and Liu]{lai2018modeling}
Lai, G., Chang, W.-C., Yang, Y., and Liu, H.
\newblock Modeling long-and short-term temporal patterns with deep neural networks.
\newblock In \emph{The 41st international ACM SIGIR conference on research \& development in information retrieval}, pp.\  95--104, 2018.

\bibitem[Li et~al.(2023)Li, Qi, Li, and Xu]{li2023revisiting}
Li, Z., Qi, S., Li, Y., and Xu, Z.
\newblock Revisiting long-term time series forecasting: An investigation on linear mapping.
\newblock \emph{arXiv preprint arXiv:2305.10721}, 2023.

\bibitem[Lieber et~al.(2024)Lieber, Lenz, Bata, Cohen, Osin, Dalmedigos, Safahi, Meirom, Belinkov, Shalev-Shwartz, et~al.]{lieber2024jamba}
Lieber, O., Lenz, B., Bata, H., Cohen, G., Osin, J., Dalmedigos, I., Safahi, E., Meirom, S., Belinkov, Y., Shalev-Shwartz, S., et~al.
\newblock Jamba: A hybrid transformer-mamba language model.
\newblock \emph{arXiv preprint arXiv:2403.19887}, 2024.

\bibitem[Liu et~al.(2023)Liu, Hu, Zhang, Wu, Wang, Ma, and Long]{liu2023itransformer}
Liu, Y., Hu, T., Zhang, H., Wu, H., Wang, S., Ma, L., and Long, M.
\newblock itransformer: Inverted transformers are effective for time series forecasting.
\newblock \emph{arXiv preprint arXiv:2310.06625}, 2023.

\bibitem[Nerlove(1971)]{nerlove1971time}
Nerlove, M.
\newblock Time series analysis, forecasting, and control., 1971.

\bibitem[Nguyen et~al.(2021)Nguyen, Turk, and McWilliams]{nguyen2021forecasting}
Nguyen, H.~M., Turk, P.~J., and McWilliams, A.~D.
\newblock Forecasting covid-19 hospital census: A multivariate time-series model based on local infection incidence.
\newblock \emph{JMIR Public Health and Surveillance}, 7\penalty0 (8):\penalty0 e28195, 2021.

\bibitem[Nie et~al.(2022)Nie, Nguyen, Sinthong, and Kalagnanam]{nie2022time}
Nie, Y., Nguyen, N.~H., Sinthong, P., and Kalagnanam, J.
\newblock A time series is worth 64 words: Long-term forecasting with transformers.
\newblock \emph{arXiv preprint arXiv:2211.14730}, 2022.

\bibitem[Vaswani(2017)]{vaswani2017attention}
Vaswani, A.
\newblock Attention is all you need.
\newblock \emph{Advances in Neural Information Processing Systems}, 2017.

\bibitem[Wang et~al.(2025)Wang, Kong, Feng, Wang, Yang, Zhao, Wang, and Zhang]{wang2025mamba}
Wang, Z., Kong, F., Feng, S., Wang, M., Yang, X., Zhao, H., Wang, D., and Zhang, Y.
\newblock Is mamba effective for time series forecasting?
\newblock \emph{Neurocomputing}, 619:\penalty0 129178, 2025.

\bibitem[Wu et~al.(2021)Wu, Xu, Wang, and Long]{wu2021autoformer}
Wu, H., Xu, J., Wang, J., and Long, M.
\newblock Autoformer: Decomposition transformers with auto-correlation for long-term series forecasting.
\newblock \emph{Advances in neural information processing systems}, 34:\penalty0 22419--22430, 2021.

\bibitem[Wu et~al.(2022)Wu, Hu, Liu, Zhou, Wang, and Long]{wu2022timesnet}
Wu, H., Hu, T., Liu, Y., Zhou, H., Wang, J., and Long, M.
\newblock Timesnet: Temporal 2d-variation modeling for general time series analysis.
\newblock \emph{arXiv preprint arXiv:2210.02186}, 2022.

\bibitem[Xu et~al.(2024)Xu, Chen, Liang, Huang, Bai, Zhao, and Shu]{xusst}
Xu, X., Chen, C., Liang, Y., Huang, B., Bai, G., Zhao, L., and Shu, K.
\newblock Sst: Multi-scale hybrid mamba-transformer experts for long-short range time series forecasting, 2024.

\bibitem[Yang et~al.(2024)Yang, Hasan, Ng, and Tarokh]{yang2024neural}
Yang, H., Hasan, A., Ng, Y., and Tarokh, V.
\newblock Neural mckean-vlasov processes: Distributional dependence in diffusion processes.
\newblock In \emph{International Conference on Artificial Intelligence and Statistics}, pp.\  262--270. PMLR, 2024.

\bibitem[Zeng et~al.(2023)Zeng, Chen, Zhang, and Xu]{zeng2023transformers}
Zeng, A., Chen, M., Zhang, L., and Xu, Q.
\newblock Are transformers effective for time series forecasting?
\newblock In \emph{Proceedings of the AAAI conference on artificial intelligence}, volume 37-9, pp.\  11121--11128, 2023.

\bibitem[Zhang \& Yan(2023)Zhang and Yan]{zhang2023crossformer}
Zhang, Y. and Yan, J.
\newblock Crossformer: Transformer utilizing cross-dimension dependency for multivariate time series forecasting.
\newblock In \emph{The eleventh international conference on learning representations}, 2023.

\bibitem[Zhou et~al.(2021)Zhou, Zhang, Peng, Zhang, Li, Xiong, and Zhang]{zhou2021informer}
Zhou, H., Zhang, S., Peng, J., Zhang, S., Li, J., Xiong, H., and Zhang, W.
\newblock Informer: Beyond efficient transformer for long sequence time-series forecasting.
\newblock In \emph{Proceedings of the AAAI conference on artificial intelligence}, volume~35, pp.\  11106--11115, 2021.

\bibitem[Zhou et~al.(2022)Zhou, Ma, Wen, Wang, Sun, and Jin]{zhou2022fedformer}
Zhou, T., Ma, Z., Wen, Q., Wang, X., Sun, L., and Jin, R.
\newblock Fedformer: Frequency enhanced decomposed transformer for long-term series forecasting.
\newblock In \emph{International conference on machine learning}, pp.\  27268--27286. PMLR, 2022.

\end{thebibliography}
\bibliographystyle{icml2025}

%%%%%%%%%%%%%%%%%%%%%%%%%%%%%%%%%%%%%%%%%%%%%%%%%%%%%%%%%%%%%%%%%%%%%%%%%%%%%%%
%%%%%%%%%%%%%%%%%%%%%%%%%%%%%%%%%%%%%%%%%%%%%%%%%%%%%%%%%%%%%%%%%%%%%%%%%%%%%%%
% APPENDIX
%%%%%%%%%%%%%%%%%%%%%%%%%%%%%%%%%%%%%%%%%%%%%%%%%%%%%%%%%%%%%%%%%%%%%%%%%%%%%%%
%%%%%%%%%%%%%%%%%%%%%%%%%%%%%%%%%%%%%%%%%%%%%%%%%%%%%%%%%%%%%%%%%%%%%%%%%%%%%%%
\newpage
\appendix
\onecolumn

\section{Algorithm}
\begin{algorithm}[H]
\caption{S2TX: State-Space Transformer With Cross Attention}
\textbf{Input:} Loss function $\mathcal{L}$, global model $g_{\phi}$, local model $f_{\psi}$, number of total training epochs $T$, dataset $D$, learning rate $\eta$, global patch length, stride, and window $PL_g$, $Str_g$, $L$, local patch length, stride, and window $PL_l$, $Str_l$, $S$. \\
\textbf{Output:} $\phi$, $\psi$.
\begin{algorithmic}[1]
\State Initialize parameters $\phi$, $\psi$.
\For{$i \gets 0$ to $T - 1$}
\State Shuffle dataset $D$.
    \For{each minibatch ($X,Y$)$\subset$ D} \comment{size}
        \State $\tilde{X}_g$, $\tilde{X}_l$ $\gets$ Patchify($X$;$PL_g$, $Str_g$,$L$), Patchify($X$;$PL_l$, $Str_l$, $S$)
        \State $Y_g \gets g_{\phi}(\tilde{X}_g)$
        \State $\hat{Y}\gets f_{\psi}(Y_g, \tilde{X}_l)$
        \State $\phi \gets \phi - \eta\nabla_{\phi}\mathcal{L}(\hat{Y},Y)$
        \State $\psi \gets \psi - \eta\nabla_{\psi}\mathcal{L}(\hat{Y},Y)$
    \EndFor
\EndFor
\State \textbf{return} $\phi$, $\psi$
\end{algorithmic}
\end{algorithm}

\section{Dataset Description}
\label{appendix:data_desc}

In this section, we describe the dataset used in our experiments in Table \ref{tab:performance}. Our experiments include 7 widely used real world multivariate time series. Table \ref{tab:datasummary} presents the number of variables and number of timesteps. 
\begin{itemize}
    \item The ETT dataset \citep{zhou2021informer} records 7 factors that related to electric transformers from July 2016 to July 2018. The ETT dataset includes 4 subsets where ETTh1 and ETTh2 are recorded hourly and ETTm1 and ETTm2 are recorded every 15 minutes. 
    \item The exchange dataset \citep{lai2018modeling} tracks the daily exchange rates of
    eight foreign countries including Australia, British, Canada,
    Switzerland, China, Japan, New Zealand, and Singapore ranging from 1990 to 2016.
    \item The weather dataset \citep{wu2021autoformer} includes 21 different meteorological features measured every 10 minutes by the Weather Station at the Max Planck Institute for Biogeochemistry.
    \item The ECL dataset \citep{lai2018modeling} records electricity consumption in kWh every 15 minutes from 2012 to 2014, for 321 clients. The data is converted to reflect hourly consumption.
\end{itemize}

\begin{table}[ht]
    \centering
    \footnotesize
    \renewcommand{\arraystretch}{1.2}
    \setlength{\tabcolsep}{8pt}
    \begin{tabular}{lccccccc}  % l = left, c = center, r = right
        \toprule
                   &ETTh1&ETTh2&ETTm1&ETTm2&Exchange& Weather & ECL\\
        \midrule
        \# Variables &7&7&7&7&8&21&321 \\
        \# Time steps & 17420& 17420 & 69680 & 69680 &  7588 & 52696 & 26304 \\
        \bottomrule
    \end{tabular}
    \caption{Table of Dataset summary including number of variables and number of time steps of each dataset. }
    \label{tab:datasummary}
\end{table}

\section{Comparison with more benchmark architectures}
\label{appendix:full_comparison}
In this section we present the full table of comparison that including two more baselines: Dlinear and FEDformer. Table \ref{tab:full_performance} is organized similarly as Table \ref{tab:performance}. 

\begin{table*}[htbp]

\centering
\renewcommand{\arraystretch}{1.1} % Adjust row height
\setlength{\tabcolsep}{3pt} % Adjust column spacing
\adjustbox{max width=\textwidth}{
\begin{tabular}{lllllllllllllllllllllll}
\toprule
 & \multicolumn{2}{c}{\textbf{S2TX}} & \multicolumn{2}{c}{\textbf{SST}} & \multicolumn{2}{c}{\textbf{S-Mamba}} & \multicolumn{2}{c}{\textbf{TimeM}} & \multicolumn{2}{c}{\textbf{iTrans}} & \multicolumn{2}{c}{\textbf{RLinear}} & \multicolumn{2}{c}{\textbf{PatchTST}} & \multicolumn{2}{c}{\textbf{CrossF}} & \multicolumn{2}{c}{\textbf{TimesNet}} & \multicolumn{2}{c}{\textbf{DLinear}} & \multicolumn{2}{c}{\textbf{FEDformer}} \\
 & \textbf{MSE} & \textbf{MAE} & \textbf{MSE} & \textbf{MAE} & \textbf{MSE} & \textbf{MAE} & \textbf{MSE} & \textbf{MAE} & \textbf{MSE} & \textbf{MAE} & \textbf{MSE} & \textbf{MAE} & \textbf{MSE} & \textbf{MAE} & \textbf{MSE} & \textbf{MAE} & \textbf{MSE} & \textbf{MAE} & \textbf{MSE} & \textbf{MAE} & \textbf{MSE} & \textbf{MAE}  \\ \midrule
\textbf{ETTh1} & & & & & & & & & & & & & & & & & & & & \\ 
96 &\textbf{0.376} &0.401&\underline{0.381} & 0.405 & 0.392 & \textbf{0.390} & 0.389 & 0.402 & 0.386 & 0.405 & 0.386 & \underline{0.395} & 0.414 & 0.419 & 0.423 & 0.448 & 0.384 & 0.402 & 0.386 & 0.400 & 0.376 & 0.419 \\ 
192 &\textbf{0.414}&\textbf{0.421}& \underline{0.430} & 0.434 & 0.449 & 0.439 & 0.435 & 0.440 & 0.441 & 0.436 & 0.437& \underline{0.424} & 0.460& 0.445 & 0.450 & 0.471 & 0.474 & 0.429 & 0.437 & 0.432 & \underline{0.420} & 0.448 \\ 
336&\textbf{0.432}&\textbf{0.435} & \underline{0.443} & \underline{0.446} & 0.467 & 0.481 & 0.450 & 0.448 & 0.487 & 0.458 & 0.479 & 0.446 & 0.501 & \underline{0.466} & 0.570 & 0.546 & 0.491 & 0.469 & 0.481 & 0.459 & 0.459 & 0.465 \\ 
720&\textbf{0.463}& \underline{0.473}& 0.502 & 0.501 & \underline{0.475} & 0.468 & 0.480 & \textbf{0.465} & 0.503 & 0.491 & 0.481 & 0.470 & 0.500 & 0.488 & 0.653 & 0.621 & 0.521 & 0.500 & 0.519 & 0.516 & 0.506 & 0.507 \\ \midrule
\textbf{ETTh2} & & & & & & & & & & & & & & & & & & & & \\ 
96&\textbf{0.279}& \underline{0.340}& 0.291 & 0.346 & 0.292 & 0.357 & 0.296 & 0.349 & 0.297 & 0.349 & \underline{0.288} & \textbf{0.338} & 0.302 & 0.348 & 0.745 & 0.584 & 0.340 & 0.374 & 0.333 & 0.387 & 0.358 & 0.397 \\ 
192&\textbf{0.362}&\underline{0.395} & \underline{0.369} & 0.397 & 0.380 & 0.402 & 0.371 & 0.400 & 0.380 & 0.400 & 0.374 & \textbf{0.390} & 0.388 & 0.400 & 0.877 & 0.656 & 0.402 & 0.414 & 0.477 & 0.476 & 0.429 & 0.439 \\ 
336&\textbf{0.337}& \textbf{0.385}& \underline{0.374} & \underline{0.414} & 0.391 & 0.420 & 0.402 & 0.449 & 0.428 & 0.432 & 0.415 & 0.426 & 0.426 & 0.433 & 1.043 & 0.731 & 0.452 & 0.452 & 0.594 & 0.541 & 0.496 & 0.487 \\ 
720&\textbf{0.395}&\textbf{0.430} & \underline{0.419} & 0.447 & 0.437 & 0.455 & 0.425 & \underline{0.438} & 0.427 & 0.445 & 0.420 & 0.440 & 0.431 & 0.446 & 1.104 & 0.763 & 0.462 & 0.468 & 0.831 & 0.657 & 0.463 & 0.474 \\ \midrule
\textbf{ETTm1} & & & & & & & & & & & & & & & & & & & & \\ 
96&\textbf{0.289}& \textbf{0.343}& \underline{0.298} & \underline{0.355} & 0.311 & 0.380 & 0.312 & 0.371 & 0.334 & 0.368 & 0.355 & 0.376 & 0.329 & 0.367 & 0.404 & 0.426 & 0.338 & 0.375 & 0.345 & 0.372 & 0.379 & 0.419 \\ 
192&\textbf{0.338}&\textbf{0.371} & \underline{0.347} & \underline{0.381} & 0.389 & 0.419 & 0.365 & 0.409 & 0.377 & 0.391 & 0.391 & 0.392 & 0.367 & 0.385 & 0.450 & 0.451 & 0.374 & 0.387 & 0.380 & 0.389 & 0.426 & 0.441 \\ 
336&\textbf{0.370}&\textbf{0.390} & \underline{0.374} & \underline{0.397} & 0.401 & 0.417 & 0.421 & 0.410 & 0.426 & 0.420 & 0.424 & 0.415 & 0.399 & 0.410 & 0.532 & 0.515 & 0.410 & 0.411 & 0.413 & 0.413 & 0.445 & 0.459 \\ 
720 &\textbf{0.423}&\textbf{0.418}& \underline{0.429} & \underline{0.428} & 0.488 & 0.476 & 0.496 & 0.437 & 0.491 & 0.459 & 0.487 & 0.450 & 0.454 & 0.439 & 0.666 & 0.589 & 0.478 & 0.450 & 0.474 & 0.453 & 0.543 & 0.490 \\ \midrule
\textbf{ETTm2} & & & & & & & & & & & & & & & & & & & & \\ 
96&\textbf{0.168}& \underline{0.260}& 0.176 & 0.264 & 0.191 & 0.301 & 0.185 & 0.290 & 0.180 & 0.264 & 0.182 & 0.265 & \underline{0.175} & \textbf{0.259} & 0.287 & 0.366 & 0.187 & 0.267 & 0.193 & 0.292 & 0.203 & 0.287 \\ 
192&\underline{0.235}&\textbf{0.298} & \textbf{0.231} & 0.303 & 0.253 & 0.312 & 0.292 & 0.309 & 0.250 & 0.309 & 0.246 & 0.304 & 0.241 & \underline{0.302} & 0.414 & 0.492 & 0.249 & 0.309 & 0.284 & 0.362 & 0.269 & 0.328 \\ 
336&\textbf{0.274}&\textbf{0.327} & \underline{0.290} & \underline{0.339} & 0.298 & 0.342 & 0.321 & 0.367 & 0.311 & 0.348 & 0.307 & 0.342 & 0.305 & 0.343 & 0.597 & 0.542 & 0.321 & 0.351 & 0.369 & 0.427 & 0.325 & 0.366 \\ 
720&\textbf{0.376}&\textbf{0.393}& \underline{0.388} & \underline{0.398} & 0.409 & 0.407 & 0.401 & 0.400 & 0.412 & 0.407 & 0.407 & 0.398 & 0.402 & 0.400 & 1.730 & 1.042 & 0.408 & 0.403 & 0.554 & 0.522 & 0.421 & 0.415 \\ \midrule
\textbf{Exchange} & & & & & & & & & & & & & & & & & & & & \\ 
96&\textbf{0.085} &\underline{0.205} &0.097 &0.222 &\underline{0.086}&0.206 & 0.089&0.208 &0.091 &0.211 & 0.088&0.209 &0.087 &\textbf{0.202} &0.095 &0.218 &0.093 & 0.211& 0.101& 0.223&0.105&0.226\\ 
192&\textbf{0.179} & \textbf{0.303}& 0.191&0.315 &0.182 &0.304 &0.184 &0.309 &0.182 & 0.303&0.188 & 0.311&\underline{0.180} &0.305 &0.193 &0.318 &0.194 &0.315 & 0.203&0.324&0.211&0.338 \\ 
336& \textbf{0.311}&\textbf{0.402} & 0.337&0.424 &0.330&0.416&0.333&0.416 &0.337 & 0.421&0.346 & 0.423& \underline{0.318}& \underline{0.407}&0.359 &0.429 &0.358 &0.433 & 0.369& 0.445&0.370&0.441\\ 
720& \textbf{0.858}&\textbf{0.696}&0.877 & 0.706&0.865 & 0.702& 0.870&\underline{0.701}&\underline{0.862} &0.703 & 0.913& 0.717&0.863 &0.703 &0.918 &0.721 &0.880 &0.719 &0.909 & 0.711&0.912&0.718\\\midrule
\textbf{Weather} & & & & & & & & & & & & & & & & & & & & \\ 
96&\textbf{0.150}& \textbf{0.199}& \underline{0.153} & \underline{0.205} & 0.169 & 0.221 & 0.174 & 0.218 & 0.174 & 0.214 & 0.192 & 0.232 & 0.177 & 0.218 & 0.158 & 0.230 & 0.172 & 0.220 & 0.196 & 0.255 & 0.217 & 0.296 \\ 
192&\textbf{0.194}&\textbf{0.242} & \underline{0.196} & \underline{0.244} & 0.205 & 0.248 & 0.200 & 0.258 & 0.221 & 0.254 & 0.240 & 0.271 & 0.225 & 0.259 & 0.206 & 0.277 & 0.219 & 0.261 & 0.237 & 0.296 & 0.276 & 0.336 \\ 
336&\underline{0.252}&\underline{0.288} & \textbf{0.246} & \textbf{0.283} & 0.288 & 0.299 & 0.280 & 0.299 & 0.278 & 0.296 & 0.292 & 0.307 & 0.278 & 0.297 & 0.272 & 0.335 & 0.280 & 0.306 & 0.283 & 0.335 & 0.339 & 0.380 \\ 
720&\textbf{0.313}&\textbf{0.333} & \underline{0.314} & \underline{0.334} & 0.335 & 0.369 & 0.352 & 0.359 & 0.358 & 0.347 & 0.364 & 0.353 & 0.354 & 0.348 & 0.398 & 0.418 & 0.365 & 0.359 & 0.345 & 0.381 & 0.403 & 0.428 \\ \midrule
\textbf{ECL} & & & & & & & & & & & & & & & & & & & & \\ 
96&\textbf{0.134}& \textbf{0.231}& \underline{0.141} & \underline{0.239} & 0.157 & 0.255 & 0.156 & 0.240 & 0.148 & 0.240 & 0.201 & 0.281 & 0.181 & 0.270 & 0.219 & 0.314 & 0.168 & 0.272 & 0.197 & 0.282 & 0.193 & 0.308 \\ 
192&\textbf{0.153}& \textbf{0.248}& \underline{0.159} & \underline{0.255} & 0.188 & 0.271 & 0.161 & 0.268 & 0.162 & 0.253 & 0.201 & 0.283 & 0.188 & 0.274 & 0.231 & 0.322 & 0.184 & 0.289 & 0.196 & 0.285 & 0.201 & 0.315 \\ 
336&\textbf{0.170}&\textbf{0.266} & \underline{0.171} & \underline{0.268} & 0.192 & 0.275 & 0.195 & 0.272 & 0.178 & 0.269 & 0.215 & 0.298 & 0.204 & 0.293 & 0.246 & 0.337 & 0.198 & 0.300 & 0.209 & 0.301 & 0.214 & 0.329 \\ 
720&\textbf{0.201}&\textbf{0.293} & \underline{0.208} & \underline{0.300} & 0.241 & 0.339 & 0.231 & 0.307 & 0.225 & 0.317 & 0.257 & 0.331 & 0.246 & 0.324 & 0.280 & 0.363 & 0.220 & 0.320 & 0.245 & 0.333 & 0.246 & 0.355 \\ \midrule
% Average& 0.304&0.349&0.315&
 \toprule
% Continue for Weather, ECL, and Traffic
\end{tabular}
}
\caption{Comprehensive comparison across various dataset with additional baselines. The \textbf{bolded} results denote the best performance, and the \underline{underlined} results indicate the second best.}
\label{tab:full_performance}
\end{table*}

\label{appendix:impl_detail}
%%%%%%%%%%%%%%%%%%%%%%%%%%%%%%%%%%%%%%%%%%%%%%%%%%%%%%%%%%%%%%%%%%%%%%%%%%%%%%%
%%%%%%%%%%%%%%%%%%%%%%%%%%%%%%%%%%%%%%%%%%%%%%%%%%%%%%%%%%%%%%%%%%%%%%%%%%%%%%%

\end{document}